\documentclass[review]{elsarticle}

\usepackage{graphicx}
\usepackage{amssymb}

\usepackage{hyperref}
\hypersetup{
	colorlinks   = true,
	citecolor    = blue
}

\usepackage{enumitem}
\usepackage{algorithm}
\usepackage{algorithmic}
\usepackage{amssymb}
\usepackage{graphicx}
\usepackage{subcaption}
\usepackage{textcomp}
\usepackage{multirow}
\usepackage{multicol}
\usepackage{makecell}
\usepackage{longtable}
\usepackage{lipsum}
\usepackage{tabularx}
\usepackage{float}
\usepackage{url}
\usepackage{dblfloatfix}
\usepackage{mathtools}
\usepackage{cuted}
\usepackage{xcolor}
\newcolumntype{L}[1]{>{\raggedright\let\newline\\\arraybackslash\hspace{0pt}}m{#1}}
\newcolumntype{C}[1]{>{\centering\let\newline\\\arraybackslash\hspace{0pt}}m{#1}}
\newcolumntype{R}[1]{>{\raggedleft\let\newline\\\arraybackslash\hspace{0pt}}m{#1}}

\usepackage{amsthm}
\newtheorem{lemma}{Lemma}

\usepackage[left=2.5cm, right=2.5cm, top=2.5cm]{geometry}

\journal{Journal of ABC}

\begin{document}

\begin{frontmatter}


\title{Accelerated learning algorithms of general fuzzy min-max neural network using a novel hyperbox selection rule}

\author[mymainaddress]{\corref{mycorrespondingauthor}Thanh Tung Khuat}\ead{thanhtung.khuat@student.uts.edu.au}

\author[mymainaddress]{Bogdan Gabrys}
\ead{bogdan.gabrys@uts.edu.au}


\cortext[mycorrespondingauthor]{Corresponding author}

\address[mymainaddress]{Advanced Analytics Institute, Faculty of Engineering and Information Technology, \\ University of Technology Sydney, NSW, Australia}



\begin{abstract}
This paper proposes a method to accelerate the training process of general fuzzy min-max neural network. The purpose is to reduce the unsuitable hyperboxes selected as the potential candidates of the expansion step of existing hyperboxes to cover a new input pattern in the online learning algorithms or candidates of the hyperbox aggregation process in the agglomerative learning algorithms. Our proposed approach is based on the mathematical formulas to form a new solution aiming to remove the hyperboxes which are certain not to satisfy expansion or aggregation conditions, and in turn decreasing the training time of learning algorithms. The efficiency of the proposed method is assessed over a number of widely used data sets. The experimental results indicated the significant decrease in training time of proposed approach for both online and agglomerative learning algorithms. Notably, the training time of the online learning algorithms is reduced from 1.2 to 12 times when using the proposed method, while the agglomerative learning algorithms are accelerated from 7 to 37 times on average.
\end{abstract}


\end{frontmatter}


\section{Introduction}
General fuzzy min-max (GFMM) neural network (GFMMNN) \cite{Gabrys2000} is a generalization framework of fuzzy min-max neural network for classification \cite{Simpson1992} and clustering \cite{Simpson1993}. The GFMM model can handle both labelled and unlabelled data as well as crisp and fuzzy input samples in a single model. One of the remarkable characteristics of the GFMMNN is that it is able to explain the predictive results based on the rule sets extracted directly or indirectly from hyperboxes \cite{Khuat2019}. This interpretable property of the GFMMNN is essential so that it can be used for high-stakes applications such as medical diagnosis, self-driving cars, and financial investment \cite{Rudin2019}. Interpretability helps fuzzy min-max neural networks to overcome the black-box drawbacks of the traditional neural networks.

There are two types of learning algorithms for the general fuzzy min-max neural network, i.e., incremental (online) learning (Onln-GFMM) \cite{Gabrys2000} and agglomerative (batch) learning \cite{Gabrys2002}. The online learning algorithm accommodates new input patterns by extending the current existing hyperboxes or creating a new hyperbox. In contrast, the agglomerative learning algorithm starts with all training samples and conducts a process of merging hyperboxes satisfying aggregation criteria to generate larger sized hyperboxes. To take advantages of the strong points of the online and agglomerative learning algorithms, a recent study proposed an improved version of online learning algorithm (IOL-GFMM) \cite{Khuat2020iol} to avoid the hyperbox contraction process which is more likely to cause the classification errors in the online learning algorithm. 

However, all of the current learning algorithms for the general fuzzy min-max neural network have the same drawback in the selection of expandable or mergeable hyperbox candidates. The creation of a new hyperbox in the online learning algorithms only occurs when all existing hyperboxes with the same class as the input patterns cannot satisfy the expansion condition to cover the new input pattern. The expansion condition is the maximum hyperbox size and the non-overlap of hyperboxes representing different classes if using the IOL-GFMM. Similarly, in the agglomerative learning algorithm, the process of hyperbox merging only terminates if all pairs of hyperbox candidates are examined with regard to the aggregation criteria but the aggregation process cannot be performed. The aggregation conditions include maximum hyperbox size, minimum similarity value, and the non-overlap of hyperboxes with different classes. The consideration of expansion or merging conditions for all hyperbox candidates leads to a waste of time. Therefore, in this study, we provide a lower bound on the membership functions and similarity measures to reduce the considered hyperbox candidates for the expansion or merging process. This method contributes to decreasing the training time of the learning algorithms.

Our contributions in this paper can be summarized as follows:

\begin{itemize}
    \item We propose and prove the lemmas to reduce significantly the considered hyperboxes for both online and batch learning algorithms for the general fuzzy min-max neural network. To the best of our knowledge, this is the first study tackling this issue.
    \item The effectiveness of the proposed method is assessed on widely used datasets. Experimental results confirmed the strong points of the method in decreasing the training time of the algorithms.
\end{itemize}

The rest of this paper is structured as follows. Section \ref{related-work} discusses several studies related to the improvements of fuzzy min-max neural networks. Section \ref{gfmm} presents an overall architecture and learning algorithms of general fuzzy min-max neural network. Section \ref{proposed_method} shows our proposed method. The experimental results and discussion are decribed in section \ref{exprt}. Section \ref{conclu} concludes the main findings.

\section{Related work} \label{related-work}
This section presents briefly several studies related to fuzzy min-max neural networks as well as its improvements. We refer the readers to a recent comprehensive survey \cite{Khuat2019} related to hyperbox-based machine learning algorithms and their applications for more details.

Since the fuzzy min-max neural network (FMNN) was proposed by Simpson \cite{Simpson1992}, there have been many studies focusing on enhancing this type of neural network. The improvements can be divided into two main directions. The first direction enhances the learning algorithms by changing the learning mechanisms, while the second research direction focuses on accelerating the learning algorithms. The direction of enhancing learning algorithms can be separated into two main groups. The first group includes the algorithms which do not allow the overlap between hyperboxes representing different classes. The second group covers the algorithms which accept the overlap between hyperboxes belonging to different classes and use a specific mechanism to handle the samples located in the overlapping regions.

General fuzzy min-max neural network \cite{Gabrys2000} is a significant improvement of the FMNN to create a single framework for both classification and clustering problems. This type of neural network does not allow the overlap occurring between hyperboxes belonging to different classes. There are two primary types of learning algorithms to build the GFMM model, which are incremental learning \cite{Gabrys2000} and agglomerative learning \cite{Gabrys2002} algorithms. As was discussed in \cite{Gabrys2004} both types of algorithms can be used as the basic building blocks within a combination of a range of algorithm independent statistical learning methods in the context of using single or multi-version GFMM for problems requiring dynamically adaptable classifiers. In a later study, Castillo and Cardenosa \cite{Castillo12} extended the incremental learning algorithm so that it can handle datasets with both numerical and categorical features. In a recent study, we improved the incremental learning of the GFMM neural network \cite{Khuat2020iol} by eliminating the contraction process, which is likely to cause the classification errors. We also proposed a new two-phase learning algorithm \cite{Khuat20} to build a learning system through many levels of abstraction. The new algorithm can reduce the number of hyperboxes generated in the training process but still maintain high classification accuracy. In addition to the GFMM model, the first group also contains other variants of the FMNN. One of these improved algorithms is an enhanced fuzzy min-max neural network (EFMNN) \cite{Mohammed15} supplementing several conditions for the overlap test. In a later study, the authors proposed the use of the K-nearest neighbor principle to select up to $K$ winning hyperboxes for the expansion process \cite{Mohammed17a}. In a recent study, Al-Sayaydeh et al. \cite{Sayaydeh19} introduced a refined fuzzy min-max neural network (RFMNN) with a new general formula for the overlap test procedure and a new hyperbox contraction operation.

The second group in the research direction on the enhancements of the learning algorithms for the original fuzzy min-max neural network introduces different methods to handle the data points falling in the overlapping regions of hyperboxes representing different classes. Bargiela et al. \cite{Bargiela04} introduced an exclusion/inclusion fuzzy min-max model. They used inclusion hyperboxes for samples of the same class and exclusion hyperboxes for samples located in the overlapping areas. In another study, Nandedkar and Biswas introduced a fuzzy min-max model with compensatory neurons (FMCN) \cite{Nandedkar07b}. The FMCN model used three types of neurons, i.e., overlap compensation, containment compensation, and classification neurons. In a later study, the authors extended the FMCN so that it can handle the input samples in both forms of crisp and hyperboxes \cite{Nandedkar09} similar to the GFMM neural network. Zhang et al. \cite{Zhang11} introduced a data-core-based fuzzy min-max model (DCFMN) removing the contraction process by adding a new type of overlapping neurons to the original structure of the FMNN. The DCFMN utilizes a new membership function considering both data core and noise. In another research, Davtalab et al. \cite{Davtalab14} proposed a multi-level fuzzy min-max neural network (MLF). Each node in the MLF model contains a subnet to handle the samples in the overlapping regions. We can see that the algorithms in this group increase the complexity of the training process. Therefore, the applicability of these algorithms for the large-sized datasets is limited.

Although there have been a lot of studies focusing on the improvement of learning algorithms for fuzzy min-max neural networks, only few studies have been devoted to accelerating the learning algorithms. Gabrys \cite{Gabrys2002b} proposed a novel method for accelerated learning by splitting a dataset into multiple exclusive subsets, learning individual GFMM models on each of the subsets in parallel and then performing a model level combination of the resulting mutliple GFMM models achieving not only significant acceleration in the training time but also a single, interpretable GFMM model as a result. In a more recent study, Ilager and Prasad \cite{Ilager17} proposed the use of MapReduce to parallelize the learning process of the original fuzzy min-max neural network for the large-sized datasets. Upasani and Om \cite{Upasani19} discussed the way of implementing the modified version of the EFMNN model using modern Graphics Processing Units (GPUs) aiming to accelerate the learning algorithm for the intrusion detection problem in the real-time. In a recent study, we introduced a method to redefine the learning process of the GFMM model using matrix operations and took advantage of GPUs to speed-up the learning algorithms for very high dimensional datasets \cite{Khuat20GPU}. However, the existing studies on accelerating the learning algorithms focus only on the use of parallel execution mechanisms. In contrast, this paper accelerates the learning algorithms by modifying the selection process of hyperbox candidates based on the proposed mathematical lemmas identifying certain relationships between key hyperparameters of the learning algorithm. This is the first work on this topic. 

\section{General fuzzy min-max neural network and learning algorithms} \label{gfmm}
\subsection{An overall architecture}
\begin{figure}
    \centering
    \includegraphics[width=0.45\linewidth]{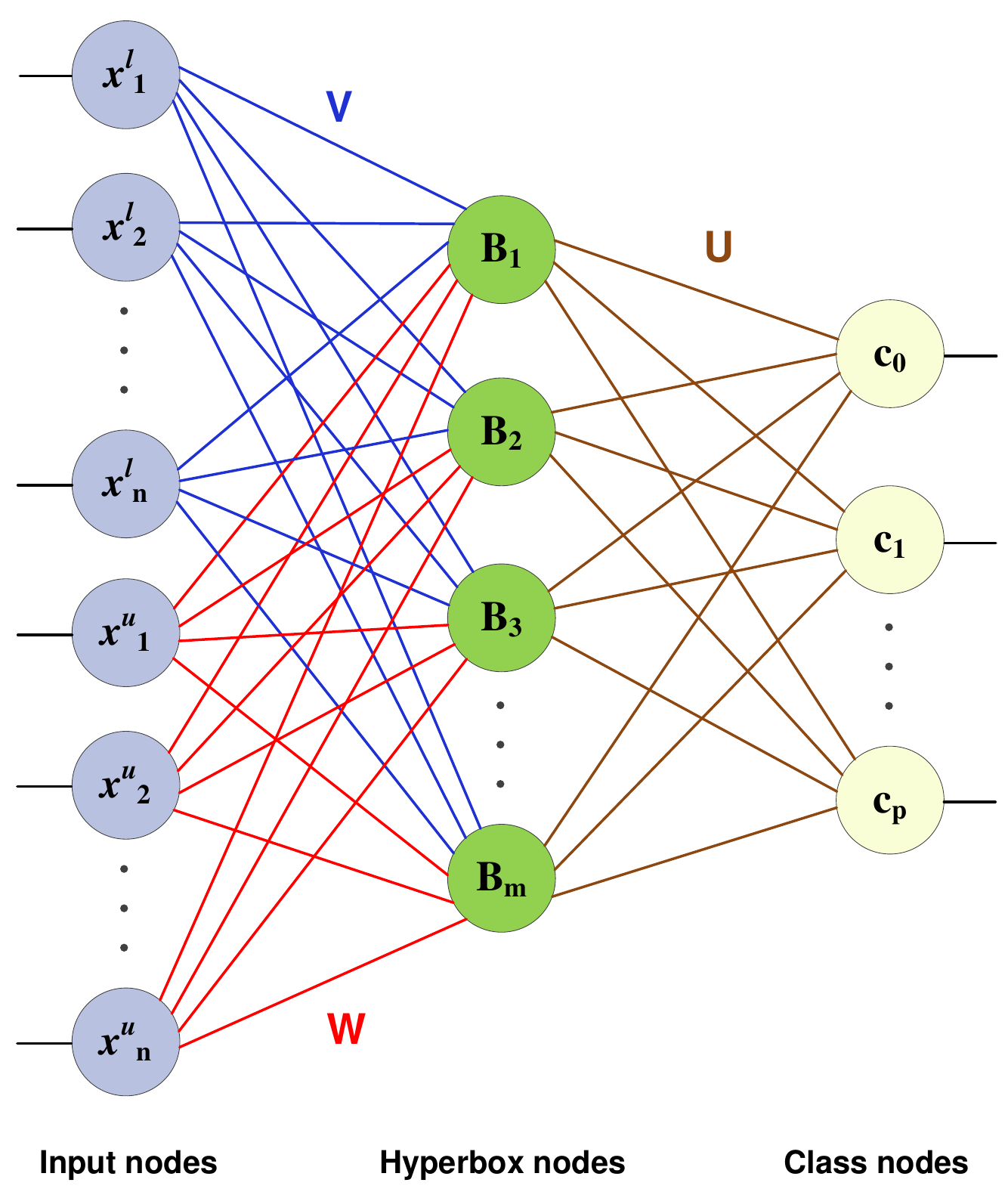}
    \caption{An architecture of general fuzzy min-max neural network}
    \label{fig1}
\end{figure}

General fuzzy min-max neural network contains three layers as shown in Fig. \ref{fig1}. The input layer includes $2n$ nodes, where $n$ is the number of dimensions of the input pattern. The first $n$ nodes are the lower bounds of the input, while the remaining $n$ nodes correspond to the upper bounds. These input nodes are connected to $m$ hyperbox nodes in the hidden layer, in which the connection weights of the lower bound input nodes are stored in a matrix $\mathbf{V}$ and the connection weights between upper bound input nodes and hyperbox nodes are saved in a matrix $\mathbf{W}$. These weights correspond to minimum points and maximum points of hyperboxes, and their values are adjusted during the training process. Each hyperbox node $B_i$ is associated with an activation function, which is also known as the membership function defined by Eq. \eqref{membership}.
\begin{equation}
\small
    b_i(X) = \min\limits_{j = 1}^n(\min([1 - f(x_j^u - w_j, \gamma_j)], [1 - f(v_j - x_j^l, \gamma_j)]))
    \label{membership}
\end{equation}
where $ f(z, \gamma) = \begin{cases} 
1, & \mbox{if } z\gamma > 1 \\
z\gamma, & \mbox{if } 0 \leq z\gamma \leq 1 \\
0, & \mbox{if } z\gamma < 0 \\
\end{cases}
$ is the ramp function, $ \gamma = [ \gamma_1,\ldots, \gamma_n ]$ is a sensitivity parameter describing the speed of decreasing of the membership function, and $X = [X^l, X^u]$ is an input pattern with lower bounds $X^l$ and upper bounds $X^u$, which are vectors with values limited in the $n$-dimensional unit hyper-cube $[0, 1]^n$.

Each hyperbox $B_i$ in the hidden layer is connected to each output (class) node $c_j$ by a binary-valued parameter $u_{ij}$ computed using Eq. \eqref{eqhidout}. There are $p + 1$ output nodes corresponding to $p$ classes, where node $c_0$ is linked to all unlabelled hyperboxes in the hidden layer. The transfer function of $ c_j $ is determined by the maximum membership value of all hyperboxes with same class as $c_j$ and is shown in Eq. \eqref{tranferfunc}.
\begin{equation}
\label{eqhidout}
    u_{ij} = \begin{cases}
        1, \quad \mbox{if hyperbox $B_i$ represents class $ c_j $} \\
        0, \quad \mbox{otherwise}
    \end{cases}
\end{equation}

\begin{equation}
    \label{tranferfunc}
    c_j = \max \limits_{i = 1}^m {b_i \cdot u_{ij}}
\end{equation}
where $m$ is the number of hyperboxes in the middle layer. The output of each class node can be a fuzzy value calculated directly from Eq. \eqref{tranferfunc} or a crisp value if the node associated with the highest membership value gets the value of one, and the others are assigned zero values \cite{Gabrys2000}.

Although the GFMM model can be applied for labelled and unlabelled datasets, this paper focuses only on the classification problems. Therefore, the learning algorithms in the next sections are presented for labeled training data. It is noted that our proposed method with the mathematical lemmas does not use the information about the class labels, so it can be applied for the unsupervised (i.e. clustering) or semi-supervised learning problems using the GFMM model as well.

\subsection{Online learning algorithm}\label{onln-gfmm}
The incremental (online) learning algorithm, proposed in \cite{Gabrys2000}, adjusts the size of existing hyperboxes or create new hyperboxes to accommodate new coming input patterns. There are three main steps in the algorithms including hyperbox expansion/creation, hyperbox overlap test, and hyperbox contraction. The pseudo code of the original online learning algorithm is given in Algorithm \ref{alg-olngfmm}.

\begin{algorithm} [!ht]
	\caption{The original online learning algorithm} \label{alg-olngfmm}
	\scriptsize{
	\begin{algorithmic} [1]
	    \REQUIRE
	    \item[]
	    \begin{itemize}
	        \item $ \theta $: The maximum hyperbox size threshold
	        \item $ \gamma $: The speed of decreasing of the membership function
	    \end{itemize}
	    \ENSURE
	    \item[]
	    A list $ \mathcal{H} $ of hyperboxes with minimum-maximum values and classes
	    \item[]
	    \STATE Initialize an empty list of hyperboxes: min-max values $ \mathcal{V} = \mathcal{W} = \varnothing $, hyperbox classes: $ \mathcal{L} = \varnothing $
	    \FOR{each input pattern $X = [X^l, X^u, l_X]$}
    	    \STATE $ n \leftarrow $ The number of dimensions of $ X $
    	    \IF{$ \mathcal{V} = \varnothing$}
    	        \STATE $ \mathcal{V} \leftarrow X^l; \quad \mathcal{W} \leftarrow X^u; \quad \mathcal{L} \leftarrow l_X $
    	    \ELSE
    	        \STATE $\mathcal{H}_1 = [\mathcal{V}_1, \mathcal{W}_1, \mathcal{L}_1] \leftarrow $ Find hyperboxes in $ \mathcal{H} = [\mathcal{V}, \mathcal{W}, \mathcal{L}] $ representing the same class as $ l_X $
        	    \STATE $ \mathcal{M} \leftarrow $ \textbf{ComputeMembershipValue}($X, \mathcal{V}_1, \mathcal{W}_1, \mathcal{L}_1$)
        	    \STATE $ \mathcal{H}_d \leftarrow $ \textbf{SortByDescending}($ \mathcal{H}_1, \mathcal{M}(\mathcal{H}_1)$)
        	    \STATE Set $ \overline{\mathcal{H}_1} \leftarrow \mathcal{H} \setminus \mathcal{H}_1 $
        	    \STATE $ flag \leftarrow false $
    	        \FOR{each $h = [V_h, W_h, l_h] \in \mathcal{H}_d$}
    	            \IF{$\mathcal{M}(h)$ = 1}
    	                \STATE $flag = true$
    	                \STATE \textbf{break}
    	            \ENDIF
    	            \IF{$\max(w_{hj}, x_{j}^u) - \min(v_{hj}, x_{j}^l) \leq \theta, \forall {j \in [1, n]} $}
    	            \STATE $ W_h^t \leftarrow \max(W_h, X^u); \quad V_h^t \leftarrow \min(V_h, X^l)$
    	            \FOR{each hyperbox $[V_i, W_i, l_i] \in \overline{H_1}$ }
    	                \STATE $isOver \leftarrow $ \textbf{OverlapTest}($V_h^t, W_h^t, V_i, W_i$)
        	            \IF{$ isOver = true $}
        	                \STATE \textbf{DoContraction}($V_h^t, W_h^t, V_i, W_i$)
        	            \ENDIF
    	            \ENDFOR
    	            \STATE $flag = true$
    	            \STATE \textbf{break}
    	            \ENDIF
    	        \ENDFOR
    	        \IF{$flag = false$}
    	            \STATE $ \mathcal{V} \leftarrow \mathcal{V} \cup X^l; \quad \mathcal{W} \leftarrow \mathcal{W} \cup X^u; \quad \mathcal{L} \leftarrow L \cup l_X $
    	        \ENDIF
    	    \ENDIF
    	\ENDFOR
		\RETURN $ \mathcal{H} = [\mathcal{V}, \mathcal{W}, \mathcal{L}] $
	\end{algorithmic}
	}
\end{algorithm}

Assuming that each input pattern is represented in the form of $X = [X^l, X^u, l_X]$, where $l_X$ is a class label and $X^l$ and $X^u$ are lower and upper bounds. The online learning algorithm first selects all existing hyperboxes with the same class as $l_X$. After that, the algorithm performs the computation of the membership values between these selected hyperboxes and the input pattern $X$, and then these membership values are sorted in a descending order (\textit{lines 8-9}). Next, the algorithm traverses in turn each hyperbox $B_i$ in the list of selected hyperboxes starting from the hyperbox with the maximum membership value to choose a hyperbox candidate aiming to expand and cover the input pattern. This process terminates when there is a hyperbox satisfying the expansion condition or the membership value is one (\textit{lines 12-28}). Otherwise, a new hyperbox will be created with the same co-ordinates and label as the input pattern (\textit{lines 29-31}). The expansion condition relates to the maximum hyperbox size threshold in each dimension as shown in Eq. \eqref{expcondi}.
\begin{equation}
    \label{expcondi}
    \max(w_{ij}, x_j^u) - \min(v_{ij}, x_j^l) \leq \theta, \quad \forall{j \in [1, n]}
\end{equation}
If this criterion is met, the hyperbox $B_i$ is extended to accommodate the input pattern $X$ using Eqs. \eqref{expand} and \eqref{expand2}.
\begin{align}
    \label{expand}
    v_{ij}^{new} &= \min(v_{ij}^{old}, x_j^l) \\
     \label{expand2}
    w_{ij}^{new} &= \max(w_{ij}^{old}, x_j^u), \quad \forall{j \in [1, n]}
\end{align}

If the expansion step of the hyperbox candidate is carried out, the extended hyperbox $B_i$ is verified for the overlap with the hyperboxes $ B_k $ representing other classes. For each dimension $j$, four following conditions are checked (initially $\delta^{old} = 1$):

\begin{itemize}
    \item $v_{ij} < v_{kj} < w_{ij} < w_{kj}: \delta^{new} = \min(w_{ij} - v_{kj}, \delta^{old}) $
    \item $ v_{kj} < v_{ij} < w_{kj} < w_{ij}: \delta^{new} = \min(w_{kj} - v_{ij}, \delta^{old}) $
    \item $ v_{ij} < v_{kj} \leq w_{kj} < w_{ij}: \delta^{new} = \min(\min(w_{kj} - v_{ij}, w_{ij} - v_{kj}), \delta^{old}) $
    \item $ v_{kj} < v_{ij} \leq w_{ij} < w_{kj}: \delta^{new} = \min(\min(w_{ij} - v_{kj}, w_{kj} - v_{ij}), \delta^{old}) $
\end{itemize}

If $ \delta^{new} < \delta^{old} $, then we set $ \Delta = j$ and $ \delta^{old} = \delta^{new} $ to show an overlapping area on the $ \Delta{th} $ dimension, and the testing procedure is repeated for the next dimension. In contrast, there is no overlap region between two considered hyperboxes, and the hyperbox contraction step will not be performed ($\Delta = -1$). If $\Delta \ne -1$, the contraction procedure is applied on the $\Delta{th}$ dimension to remove the overlapping area between two hyperboxes. The overlapping region is eliminated by tuning the value of the dimension with the smallest overlap. If $ \Delta > 0$, this dimension is adjusted according to the four following cases:

Case 1: $ v_{i\Delta} < v_{k\Delta} < w_{i\Delta} < w_{k\Delta}: w_{i\Delta}^{new} = v_{k\Delta}^{new} = (w_{i\Delta}^{old} + v_{k\Delta}^{old}) / 2 $

Case 2: $ v_{k\Delta} < v_{i\Delta} < w_{k\Delta} < w_{i\Delta}:
        w_{k\Delta}^{new} = v_{j\Delta}^{new} = (w_{k\Delta}^{old} + v_{j\Delta}^{old}) / 2 $

Case 3: $ v_{i\Delta} < v_{k\Delta} \leq w_{k\Delta} < w_{i\Delta}: $
        \begin{align*}
        & v_{i\Delta}^{new} = w_{k\Delta}^{old}, \quad \mbox{if } w_{k\Delta} - v_{i\Delta} \leq w_{i\Delta} - v_{k\Delta} \\
        & w_{i\Delta}^{new} = v_{k\Delta}^{old}, \quad \mbox{if } w_{k\Delta} - v_{i\Delta} > w_{i\Delta} - v_{k\Delta}
    \end{align*}
        
Case 4: $ v_{k\Delta} < v_{i\Delta} \leq w_{i\Delta} < w_{k\Delta}: $
        \begin{align*}
        & w_{k\Delta}^{new} = v_{i\Delta}^{old}, \quad \mbox{if } w_{k\Delta} - v_{i\Delta} \leq w_{i\Delta} - v_{k\Delta} \\
        & v_{k\Delta}^{new} = w_{i\Delta}^{old}, \quad \mbox{if } w_{k\Delta} - v_{i\Delta} > w_{i\Delta} - v_{k\Delta}
    \end{align*}

\paragraph{Time complexity of the Onln-GFMM algorithm} In terms of time complexity, assuming that there are $N$ training samples with $n$ features, the algorithm will first traverse each input sample and find a list of $\mathcal{K}$ hyperboxes with the same class as the input sample. The time complexity for this operation is constant if we use the hashtable technique. The membership computation must check all $n$ dimensions of $\mathcal{K}$ hyperboxes, so the time complexity is $\mathcal{O}(\mathcal{K} n)$. We obtain $\mathcal{K}$ membership values for each input, therefore, the time complexity of the sorting operation is $\mathcal{O}(\mathcal{K} \log \mathcal{K})$. Let $\mathcal{R}$ be the number of hyperboxes representing classes different from the input class in the current iteration, we need $\mathcal{O(\mathcal{R})}$ to collect these $\mathcal{R}$ hyperboxes. In the worst-case, we have to traverse over all $\mathcal{K}$ selected hyperboxes to find the expandable hyperbox (\textit{line 12}). For each hyperbox candidate, the checking of the expansion condition through $n$ dimensions requires $\mathcal{O}(n)$. The overlap test between the hyperbox candidate and $\mathcal{R}$ hyperboxes belonging to other classes requires $\mathcal{O}(\mathcal{R} n)$ for time complexity. We need only constant time for the process of contraction or generating a new hyperbox. Hence, the time complexity from \textit{line 12} to \textit{line 28} in the worst-case is $\mathcal{O}(\mathcal{K} \mathcal{R} n)$. Finally, let $\overline{\mathcal{K}}$ be the average number of hyperbox candidates for each iteration, $\overline{\mathcal{R}}$ be the average number of hyperboxes representing classes different from the input class over $N$ training samples, the time complexity of learning algorithm \ref{alg-olngfmm} in the worst-case is $\mathcal{O}(N \cdot \overline{\mathcal{K}} \cdot \overline{\mathcal{R}} \cdot n)$.

\subsection{Agglomerative learning algorithm}\label{agglo-sec}
The online learning algorithm creates or adjusts the size of hyperboxes whenever an input sample comes in the network. Therefore, its performance depends on the data presentation order. In \cite{Gabrys2002}, Gabrys proposed an agglomerative learning algorithm based on the full similarity matrix (AGGLO-SM) to reduce the impact of the data presentation order on the accuracy of the learning algorithm. In contrast to the online learning algorithm, the AGGLO-SM algorithm for the classification problems starts with all of the training samples. The idea is to merge hyperboxes with the same class, possessing the similarity values larger than a given threshold, and not generating the overlapping areas with existing hyperboxes representing other classes. The main steps of the AGGLO-SM algorithm are shown in Algorithm \ref{alg-agglofullgfmm}.

\begin{algorithm} [!ht]
	\caption{The agglomerative algorithm with full similarity matrix - AGGLO-SM} \label{alg-agglofullgfmm}
	\scriptsize{
	\begin{algorithmic} [1]
	    \REQUIRE
	    \item[]
	    \begin{itemize}
	        \item $ \mathbf{X} = [\mathbf{X}^l, \mathbf{X}^u]$: A list of training features
	        \item $ \mathbf{L}$: A vector of pattern classes
	        \item $ \theta $: The maximum hyperbox size threshold
	        \item $ \gamma $: The speed of decreasing of the membership function
	    \end{itemize}
	    \ENSURE
	    \item[]
	    A list $ \mathcal{H} $ of hyperboxes with minimum-maximum values and classes
	    \item[]
	    \STATE Initialize a list of hyperboxes: min-max values $ \mathcal{V} = \mathbf{X}^l, \mathcal{W} = \mathbf{X}^u $, hyperbox classes: $ \mathcal{L} = \mathbf{L} $
	    \STATE $loop \leftarrow true$; $ n \leftarrow $ the number of features of $\mathbf{X}$
	    \STATE $ S \leftarrow $ \textbf{ComputeSimilarityValPairWithinEachClass}($\mathcal{V}, \mathcal{W}, \mathcal{L}$)
	    \WHILE{$loop = true$}
    	    \STATE $ loop \leftarrow false$
    	    \STATE $ S \leftarrow S \setminus \{ s \in S | s < \sigma \}$
        	\STATE $ I, K, S \leftarrow $ \textbf{SortByDescending}($ S, \mathcal{V}, \mathcal{W}, \mathcal{L}$)
    	    \FOR{each $[i, k, s] \in [I, K, S]$}
    	       \IF{$\max(w_{ij}, w_{kj}) - \min(v_{ij}, v_{kj}) \leq \theta, \forall {j \in [1, n]} $}
    	            \STATE $ W_t \leftarrow \max(W_i, W_k); \quad V_t \leftarrow \min(V_i, V_k)$
    	            \STATE $\overline{H_1} \leftarrow $ A list of hyperboxes with classes different from $l_i \in \mathcal{L}$
    	           \STATE $isOver \leftarrow $ \textbf{IsOverlap}($V_t, W_t, \overline{\mathcal{H}_1}$)
        	       \IF{$ isOver = false $}
        	           \STATE $loop \leftarrow true$
        	           \STATE $V_i \leftarrow V_t; \quad W_i \leftarrow W_t$
        	           \STATE $\mathcal{V} \leftarrow \mathcal{V} \setminus V_k; \quad \mathcal{W} \leftarrow \mathcal{W} \setminus W_k; \quad \mathcal{L} \leftarrow \mathcal{L} \setminus L_k$
        	           \STATE $ S \leftarrow $ \textbf{UpdateSimilarityMatrix}($\mathcal{V}, \mathcal{W}, \mathcal{L}$)
        	           \STATE \textbf{break}
        	       \ENDIF
    	        \ENDIF
    	        \ENDFOR
    	\ENDWHILE
		\RETURN $ \mathcal{H} = [\mathcal{V}, \mathcal{W}, \mathcal{L}] $
	\end{algorithmic}
	}
\end{algorithm}

Firstly, the algorithm initializes a matrix $\mathcal{V}$ of minimum points and a matrix $\mathcal{W}$ of maximum points using the lower bounds \textbf{$X^l$} and upper bounds \textbf{$X^u$} of all training samples. Next, the algorithm performs a repeated training process of aggregating hyperboxes starting from the computation of a similarity matrix of hyperboxes for each class. There are three measures possible to be used to find the similarity value of each pair of hyperboxes $ B_i $ and $ B_k $ as follows:

\begin{itemize}
    \item The first similarity measure is based on maximum points or minimum points of two hyperboxes. For simplifying, we call this measure as ``middle distance" in this paper, though the similarity measures are not distance measures: \\ $ s_{ik} = s(B_i, B_k) = \min \limits_{j = 1}^{n}(\min(1 - f(w_{kj} - w_{ij}, \gamma_j), 1 - f(v_{ij} - v_{kj}, \gamma_j))) $ \\
    It can be seen that $ s_{ik} \neq s_{ki} $, thus the similarity value between $ B_i $ and $ B_k $ may receive the minimum or maximum value between $ s_{ik} $ and $ s_{ki} $. If the maximum value is used, this measure is called ``mid-max distance"; otherwise, the name ``mid-min distance" is used.
    \item The second similarity measure employs the smallest gap between two hyperboxes $B_i$ and $B_k$, namely ``shortest distance" in this paper:
    \begin{equation}
        \label{simi_shortest}
        \widetilde{s}_{ik} = \widetilde{s}(B_i, B_k) = \min \limits_{j = 1}^{n}(\min(1 - f(v_{kj} - w_{ij}, \gamma_j), 1 - f(v_{ij} - w_{kj}, \gamma_j)))
    \end{equation}
    
    \item The third similarity measure is based on the longest possible distance between two hyperboxes $B_i$ and $B_k$, called ``longest distance" for short, defined as follows: \\
    $ \widehat{s}_{ik} = \widehat{s}(B_i, B_k) = \min \limits_{j = 1}^{n}(\min(1 - f(w_{kj} - v_{ij}, \gamma_j), 1 - f(w_{ij} - v_{kj}, \gamma_j))) $ \\
    We can observe that both $ \widetilde{s}_{ik} $ and $ \widehat{s}_{ik}  $ possess the symmetrical property.
\end{itemize}

From the similarity matrix of hyperboxes with the same class, the algorithm will merge sequentially hyperboxes by seeking for a pair of hyperboxes with the maximum similarity value. It is noted that the algorithm only considers pairs of hyperboxes with similarity values larger than or equal to a minimum similarity threshold ($ \sigma $): $ s_{ih} \geq \sigma $ (\textit{line 6} - Algorithm \ref{alg-agglofullgfmm}). Assuming that these two hyperboxes are $B_i$ and $B_k$, the following conditions will be checked before aggregating:

\begin{enumerate}[label=(\alph*)]
    \item Maximum hyperbox size: \\
    $ \max(w_{ij}, w_{kj}) - \min(v_{ij}, v_{kj}) \leq \theta, \quad \forall j \in [1, n] $
    \item Overlap test. Newly aggregated hyperbox from $ B_i $ and $ B_k $ does not overlap with any existing hyperboxes belonging to other classes. The overlap checking conditions between two hyperboxes are shown in subsection \ref{onln-gfmm}. If any overlapping area exists, another pair of hyperboxes is selected.
\end{enumerate}

If all above constraints are met, the hyperbox aggregation process is carried out as follows:

\begin{enumerate}[label=(\alph*)]
    \item Updating the coordinates of $ B_i $ so that $ B_i $ represents the coordinates of the merged hyperbox (\textit{line 15}).
    \item Removing $ B_k $ from the current set of hyperboxes (\textit{line 16}) and update the similarity matrix (\textit{line 17}).
\end{enumerate}

This training process is iterated until there are no pairs of hyperboxes to aggregate.
\paragraph{Time complexity of the AGGLO-SM algorithm} The time complexity of the computation of similarity values at \textit{line 3} in Algorithm \ref{alg-agglofullgfmm} is $\mathcal{O}(N^2n)$, where $N$ is the number of training samples and $n$ is the number of features. With $N$ training samples, we obtain a maximum of $N(N - 1)/2$ hyperbox pairs. In the worst-case, we have to loop through all pairs of hyperboxes. At each iteration, let $\mathcal{Y}$ be the number of existing hyperboxes, the filtering step of hyperbox pairs satisfying the minimum similarity value (\textit{line 6}) requires $\mathcal{O}(\mathcal{Y}^2)$ for time complexity. Assuming we obtain $\mathcal{Z}$ pairs of hyperbox candidates for the aggregation process, the complexity of the sorting step is $\mathcal{O}(\mathcal{Z} \log \mathcal{Z})$. In the worst-case, we need to check all of these $\mathcal{Z}$ candidate pairs for the aggregation process. For each pair, the checking for the maximum hyperbox size condition requires $\mathcal{O}(n)$. The process of collecting hyperboxes representing classes different from the newly aggregated hyperbox takes $\mathcal{O}(\mathcal{Y})$. The overlap test between the newly aggregated hyperbox and existing hyperboxes requires $\mathcal{O}(\mathcal{Y}n)$. The update step of the similarity matrix takes $\mathcal{O}(\mathcal{Y}n)$. Therefore, in the worst-case, the process of aggregating a pair of hyperboxes (\textit{lines 6-21} in Algorithm \ref{alg-agglofullgfmm}) requires $\mathcal{O}(\mathcal{Z} \mathcal{Y} n + \mathcal{Y}^2)$ for time complexity. As a result, let $\overline{\mathcal{Z}}$ be the average number of pairs of hyperbox candidates considered during the training process and $\overline{\mathcal{Y}}$ be the number of existing hyperboxes in each iteration, the time complexity for the AGGLO-SM algorithm is $\mathcal{O}(N^2 \cdot (\overline{\mathcal{Z}} \cdot \overline{\mathcal{Y}} \cdot n + \overline{\mathcal{Y}}^2))$ in the worst-case.

Training process of the AGGLO-SM algorithm takes a very long time to complete, especially for massive datasets, due to the fact that we need to compute and sort the similarity matrix for all pairs of hyperboxes. To lower the training time of the AGGLO-SM algorithm, Gabrys \cite{Gabrys2002} introduced the second agglomerative algorithm (AGGLO-2) removing the usage of the full similarity matrix when selecting and merging hyperboxes. The main steps of the AGGLO-2 are shown in Algorithm \ref{alg-agglo2}.
 
 The AGGLO-2 algorithm traverses and selects in turn each hyperbox in the current list of hyperboxes to perform the hyperbox merging process. For the first selected hyperbox candidate $ B_i $, the similarity values of $ B_i $ and the remaining hyperboxes with the same class as $B_i$ are computed and sorted (\textit{lines 7-9} in Algorithm \ref{alg-agglo2}). The hyperbox $ B_k $ with the highest similarity value is chosen as the second candidate for the aggregation. The aggregation constraints of $B_i$ and $B_k$ are the same as in the AGGLO-SM algorithm. If the aggregation conditions are met, we update the coordinates of $B_i$ so that it shows the aggregated hyperbox and remove $B_k$ of the list of current hyperboxes. If $B_i$ and $B_k$ do not meet the constraints, the hyperbox with the second largest similarity value is selected, and the above checking and merging steps are repeated until the agglomeration happens. The learning algorithm stops when no pair of hyperboxes can be aggregated (the variable $loop = false$ in Algorithm \ref{alg-agglo2}).

\begin{algorithm} [!ht]
	\caption{The agglomerative algorithm version two - AGGLO-2} \label{alg-agglo2}
	\scriptsize{
	\begin{algorithmic} [1]
	    \REQUIRE
	    \item[]
	    \begin{itemize}
	        \item $ \mathbf{X} = [\mathbf{X}^l, \mathbf{X}^u]$: A list of training features
	        \item $ \mathbf{L}$: A vector of pattern classes
	        \item $ \theta $: The maximum hyperbox size threshold
	        \item $ \gamma $: The speed of decreasing of the membership function
	    \end{itemize}
	    \ENSURE
	    \item[]
	    A list $ \mathcal{H} $ of hyperboxes with minimum-maximum values and classes
	    \item[]
	    \STATE Initialize a list of hyperboxes: min-max values $ \mathcal{V} = \mathbf{X}^l, \mathcal{W} = \mathbf{X}^u $, hyperbox classes: $ \mathcal{L} = \mathbf{L} $
	    \STATE $loop \leftarrow true$; $ n \leftarrow $ the number of features of $\mathbf{X}$
	    \WHILE{$loop = true$}
    	    \STATE $ loop \leftarrow false; \quad i \leftarrow 1$
    	    \WHILE{$i \leq |\mathcal{L}|$}
    	    \STATE $\mathcal{H}_1 = [\mathcal{V}_1, \mathcal{W}_1, \mathcal{L}_1] \leftarrow $ Find hyperboxes in $[\mathcal{V}, \mathcal{W}, \mathcal{L}] $ representing the same class as $l_i \in \mathcal{L}$
    	    \STATE $ S \leftarrow $ \textbf{ComputeSimilarityValPair}($V_i, W_i, \mathcal{H}_1$)
    	    \STATE $ S \leftarrow S \setminus \{ s \in S | s < \sigma \}$
        	\STATE $ K, S \leftarrow $ \textbf{SortByDescending}($ S, \mathcal{V}_1, \mathcal{W}_1, \mathcal{L}_1$)
    	    \FOR{each $[k, s] \in [K, S]$}
    	       \IF{$\max(w_{ij}, w_{kj}) - \min(v_{ij}, v_{kj}) \leq \theta, \forall {j \in [1, n]} $}
    	            \STATE $ W_t \leftarrow \max(W_i, W_k); \quad V_t \leftarrow \min(V_i, V_k)$
    	            \STATE $\overline{H_1} \leftarrow $ A list of hyperboxes with classes different from $l_i$
    	           \STATE $isOver \leftarrow $ \textbf{IsOverlap}($V_t, W_t, \overline{\mathcal{H}_1}$)
        	       \IF{$ isOver = false $}
        	           \STATE $loop \leftarrow true$
        	           \STATE $V_i \leftarrow V_t; \quad W_i \leftarrow W_t$
        	           \STATE $\mathcal{V} \leftarrow \mathcal{V} \setminus V_k; \quad \mathcal{W} \leftarrow \mathcal{W} \setminus W_k; \quad \mathcal{L} \leftarrow \mathcal{L} \setminus L_k$
        	           \IF{$i > k$}
        	                \STATE $i \leftarrow i - 1$
        	           \ENDIF
        	           \STATE \textbf{break}
        	       \ENDIF
    	        \ENDIF
    	        \ENDFOR
    	        \STATE $i \leftarrow i + 1$
    	   \ENDWHILE
    	\ENDWHILE
		\RETURN $ \mathcal{H} = [\mathcal{V}, \mathcal{W}, \mathcal{L}] $
	\end{algorithmic}
	}
\end{algorithm}

\paragraph{Time complexity of the AGGLO-2 algorithm} Assuming that we have $N$ training samples with $n$ features, in the worst-case, the training process has to loop through all $N$ initial hyperboxes. In each iteration, the process of finding the hyperboxes representing the same class as the considered hyperbox takes constant time if using the hashtable. Let $\mathcal{K}$ be the number of hyperboxes with the same class as the considered hyperbox in the current iteration, the computation step of similarity values takes $\mathcal{O}(\mathcal{K}n)$. The filtering step of hyperbox candidates satisfying the minimum similarity value takes $\mathcal{O}(\mathcal{K})$. Let $\mathcal{Z}$ be the number of pairs of hyperbox candidate for the aggregation process, the sorting step requires $\mathcal{O}(\mathcal{Z \log Z})$. In the worst-case, the aggregation process needs to loop through all $\mathcal{Z}$ pairs of hyperbox candidates. For each candidate pair, the checking step of the maximum hyperbox size condition requires $\mathcal{O}(n)$. Let $\mathcal{R}$ be the number of hyperboxes representing classes different from the considered pair of hyperbox candidate, we need to take $\mathcal{O}(\mathcal{R})$ to find these hyperboxes. The overlap test step (\textit{line 14} in Algorithm \ref{alg-agglo2}) requires $\mathcal{O}(\mathcal{R}n)$. The operation of removing a merged hyperbox requires $\mathcal{O}(\mathcal{K})$ but this step only occurs once among $\mathcal{Z}$ pairs of hyperbox candidates. As a result, the time complexity for the aggregation process (\textit{lines 6-26}) is $\mathcal{O}(\mathcal{Z}\mathcal{R}n)$. In summary, let $\overline{\mathcal{Z}}$ be the average number of pairs of hyperbox candidates and $\overline{\mathcal{R}}$ be the average number of hyperboxes belonging to classes different from the class of the considered pair of hyperbox, the time complexity of the AGGLO-2 algorithm is $\mathcal{O}(N \cdot \overline{\mathcal{Z}} \cdot \overline{\mathcal{R}} \cdot n)$ in the worst-case.

\subsection{An improved online learning algorithm}
The improved online learning algorithm (IOL-GFMM) is proposed in \cite{Khuat2020iol} to tackle the disadvantages of the original online learning algorithms related to the hyperbox contraction and equal membership value. Similarly to the agglomerative learning algorithm, the IOL-GFMM algorithm prevents the overlap between the expanded hyperbox and any existing hyperboxes representing other classes, so it removes the contraction step from the learning algorithm. The learning process contains only two main steps, i.e., hyperbox expansion/creation and overlap test. The details of the IOL-GFMM are given in Algorithm \ref{alg-iolgfmm}.

\begin{algorithm} [!ht]
	\caption{The IOL-GFMM algorithm} \label{alg-iolgfmm}
	\scriptsize{
	\begin{algorithmic} [1]
	    \STATE Lines 1-11 from the \textbf{Algorithm} \ref{alg-olngfmm}
    	        \FOR{each $h = [V_h, W_h, l_h] \in \mathcal{H}_d$}
    	            \IF{$\mathcal{M}(h)$ = 1}
    	                \STATE $flag = true$
    	                \STATE Increase the number of samples contained in $h$
    	                \STATE \textbf{break}
    	            \ENDIF
    	            \IF{$\max(w_{hj}, x_{j}^u) - \min(v_{hj}, x_{j}^l) \leq \theta, \forall {j \in [1, n]} $}
    	            \STATE $ W_h^t \leftarrow \max(W_h, X^u); \quad V_h^t \leftarrow \min(V_h, X^l)$
    	            \STATE $isOver \leftarrow $ \textbf{IsOverlap}($W_h^t, V_h^t, \overline{\mathcal{H}_1}$)
    	            \IF{$ isOver = false$}
    	                \STATE $ V_h \leftarrow V_h^t; \quad W_h \leftarrow W_h^t$
    	                \STATE $ flag \leftarrow true$
    	                \STATE Increase the number of samples contained in $h$
    	                \STATE \textbf{break}
    	            \ENDIF
    	            \ENDIF
    	        \ENDFOR
    	        \STATE Lines 29-34 from the \textbf{Algorithm} \ref{alg-olngfmm}.
	\end{algorithmic}
	}
\end{algorithm}

\paragraph{\textit{Expansion of hyperboxes}} When an input sample $ X = [X^l, X^u, l_X] $ comes to the network, the algorithm first filters all existing hyperboxes representing the same class as $ l_X $. Next, the membership values between $X$ and all selected hyperboxes are computed and sorted in a descending order. If the maximum membership value is one, the learning process continues with another input sample. Otherwise, the algorithm traverses in turn each hyperbox starting from the hyperbox with the maximum membership value to verify the expansion criteria. If all constraints are satisfied, the size of the selected hyperbox and the number of pattern included in that hyperbox are updated and the learning process continues with next input samples (\textit{lines 2-18}). If none of the hyperbox candidates satisfies the conditions, a new hyperbox is generated to accommodate the input pattern and added to the current set of hyperboxes. Two expansion criteria include the maximum hyperbox size shown in Eq. \eqref{expcondi} and overlap. If the maximum hyperbox size constraint is satisfied, then the non-overlapping condition is verified as follows:

\paragraph{\textit{Overlap test}} The overlap test is performed between the newly extended hyperbox and the hyperboxes belonging to other classes. If there is any overlapping regions occurring, the next hyperbox candidate is considered. Otherwise, the chosen hyperbox is updated by the new coordinates of the expanded hyperbox, and the learning steps continue with another input sample. The overlap test between each pair of hyperboxes is conducted in the same way as the steps shown in subsection \ref{onln-gfmm}.

\paragraph{\textit{Classification phase}}
In the classification phase, when an unseen sample $X$ comes to the network, the membership degrees between $X$ and all existing hyperboxes in the model are calculated. The input $X$ will be classified to the class of the hyperbox with the maximum membership value. In the original online learning algorithm, if there are at least two hyperboxes with the same maximum membership value but different classes, the algorithm will select the predicted class randomly. In contrast, in the IOL-GFMM algorithm, if many hyperboxes belonging to $ K $ different classes output the same maximum membership degree ($ b_{win}$), we need to deploy an additional criterion to specify the suitable class for $X$. If $ b_{win} = 1$ and $\exists i: n_i = 1$, then the class of $X$ is the class of $B_i$, where $n_i$ is the number of patterns included in the hyperbox $ B_i$. Otherwise, the predicted class of $X$ is the class $c_k$ with the highest value of $\mathcal{P}(c_k|X) $ defined by:
\begin{equation}
    \label{probcard}
    \mathcal{P}(c_k|X) = \cfrac{\sum_{j \in \mathcal{I}_{win}^k} n_j \cdot b_j}{\sum_{i \in \mathcal{I}_{win}} n_i \cdot b_i }
\end{equation}
where $k \in [1, K]$ and $ \mathcal{I}_{win} = \{ i, \mbox{if } b_i = b_{win} \}$ contains the indexes of all hyperbox with the same maximum membership value, $ I_{win}^k = \{ j, \mbox{if } class(B_j) = c_k \mbox{ and } b_j = b_{win} \}$ is a subset of $ I_{win} $ created by indexes of the $k^{th}$ class.

\paragraph{Time complexity of the IOL-GFMM algorithm} We can see that the IOL-GFMM is different from the original online learning algorithm of the GFMM neural network in the contraction step only. Therefore, the time complexity of the IOL-GFMM algorithm is the same as that of the Onln-GFMM algorithm in the worst-case, i.e., $\mathcal{O}(N \cdot \overline{\mathcal{K}} \cdot \overline{\mathcal{R}} \cdot n)$, where $N$ is the number of training samples, $n$ is the number of features, $\overline{\mathcal{K}}$ is the average number of expandable hyperbox candidates considered during the training process, and $\overline{\mathcal{R}}$ is the average number of hyperboxes representing classes different from the class of the current training sample.

\section{Proposed method}\label{proposed_method}
In order to alleviate the computational issues, we present an approach to training algorithms that drastically reduces the number of considered expandable hyperbox candidates by omitting hyperboxes certain not to satisfy the expansion or aggregation conditions.

\subsection{Accelerated online learning algorithms}
It is observed that in online learning algorithms of GFMMNN, a new hyperbox is only created to cover the input pattern if all hyperbox candidates cannot satisfy the conditions to be expanded for accommodating the new input pattern. However, in the current versions of the online learning algorithms, there is no way to reduce the considered hyperbox candidates. In this paper, therefore, we provide a lemma to narrow down the expandable hyperboxes during the training process. This solution is given in Lemma \ref{lemma1}. 

\begin{lemma} \label{lemma1}
When finding the candidates of expandable hyperboxes to cover an input pattern $X \in [0, 1]^n$, we only need to consider the hyperboxes ($h$) with the same class as $X$ and having a membership degree ($b_h(X)$) to the new input pattern satisfying: $ b_h(X) \geq 1 - \theta \cdot \gamma_{max}$, where $ \gamma_{max} = \max \limits_{j=1}^n (\gamma_j); \gamma_j > 0$, and $\theta$ is the maximum hyperbox size. If the input pattern $X$ is in the form of a hyperbox, its size must be below $\theta$ in all dimensions.
\end{lemma}

The proof of Lemma \ref{lemma1} can be found in the \ref{appendix-a}. This lemma shows the relationship between the membership function and the maximum hyperbox size parameter if we keep the sensitivity parameter $\gamma$ fixed. By using this lemma, we can reduce the number of hyperbox candidates considered for the expansion step based on their membership values. We can modify the Algorithm \ref{alg-olngfmm} into Algorithm \ref{alg-acceolngfmm} to accelerate the original online learning algorithm of GFMM model. The only change in this algorithm is that we use the proposed lemma to limit the number of hyperboxes considered for each input pattern. The other steps are the same as in the original version.

\begin{algorithm} [!ht]
	\caption{The accelerated original online learning algorithm} \label{alg-acceolngfmm}
	\scriptsize{
	\begin{algorithmic} [1]
	    \STATE Lines 1-8 from the \textbf{Algorithm} \ref{alg-olngfmm}
	    \STATE $\mathcal{H}_{s} \leftarrow \{h : h \in \mathcal{H}_1, \mathcal{M}(h) \ge 1 - \theta \cdot \gamma_{max}\}$
	    \STATE $ \mathcal{H}_d \leftarrow $ \textbf{SortByDescending}($ \mathcal{H}_s, \mathcal{M}(\mathcal{H}_s)$)
	    \STATE Lines 10-34 from \textbf{Algorithm} \ref{alg-olngfmm}
	\end{algorithmic}
	}
\end{algorithm}

Similarly, we can also change the steps of the IOL-GFMM shown in Algorithm \ref{alg-iolgfmm} to Algorithm \ref{alg-acceiolgfmm} to accelerate the IOL-GFMM procedure. With the Lemma \ref{lemma1}, we can reduce the extendable hyperbox candidates to cover the new input pattern. The remaining operations are the same as the original version of the IOL-GFMM algorithm. 

\begin{algorithm} [!ht]
	\caption{The accelerated IOL-GFMM algorithm} \label{alg-acceiolgfmm}
	\scriptsize{
	\begin{algorithmic} [1]
	    \STATE Lines 1-8 from the \textbf{Algorithm} \ref{alg-olngfmm}
	    \STATE $\mathcal{H}_{s} \leftarrow \{h : h \in \mathcal{H}_1, \mathcal{M}(h) \ge 1 - \theta \cdot \gamma_{max}\}$
	    \STATE $ \mathcal{H}_d \leftarrow $ \textbf{SortByDescending}($ \mathcal{H}_s, \mathcal{M}(\mathcal{H}_s)$)
	    \STATE Set $ \overline{\mathcal{H}_1} \leftarrow \mathcal{H} \setminus \mathcal{H}_1 $
	    \STATE $ flag \leftarrow false $
	    \STATE Lines 2-19 from \textbf{Algorithm} \ref{alg-iolgfmm}
	\end{algorithmic}
	}
\end{algorithm}

As will be illustrated in the experimental section, these changes to the algorithms and the use of the proved Lemma \ref{lemma1} resulted in from 2 to 3.5 times reduction of learning time on average.

\paragraph{Time complexity of the accelerated online learning algorithms} It is easily observed that the accelerated online learning algorithms are different from the original version in the reduction of the hyperbox candidates considered during the training process. Therefore, the time complexity of these algorithms are similar to the original versions, i.e., $\mathcal{O}(N \cdot \overline{\mathcal{K}}_1 \cdot \overline{\mathcal{R}} \cdot n)$, where $\overline{\mathcal{K}}_1$ is the average number of expandable hyperbox candidates considered during the training process, the meaning of other parameters is the same as in the original versions. The accelerated learning algorithms run faster than the original versions because of $\overline{\mathcal{K}}_1 < \overline{\mathcal{K}}$.

\subsection{Accelerated agglomerative learning algorithms}
In the original agglomerative learning algorithms, the hyperbox aggregation process considers all pairs of hyperboxes for which their similarity values are larger than or equal to a given minimum similarity threshold. If this threshold is set too small, then there might be many candidates considered, and so the training process can be long. However, when the minimum similarity condition is met and two hyperboxes could be merged, the newly aggregated hyperbox has to still be checked for the maximum hyperbox size constraint. In this paper, we will show the dependency of the similarity value with the maximum hyperbox size parameter. Based on this identified relationship, we can remove immediately the pairs of hyperboxes for which the hyperbox aggregated from these candidates cannot, with absolute certainty, satisfy the maximum hyperbox size condition. The details of the proposed method are described in Lemma \ref{lemma2}.

\begin{lemma} \label{lemma2}
Regardless of the similarity measure used, the hyperbox aggregation process only considers pairs of hyperbox candidates that their similarity values satisfy the following condition: $s(B_i, B_k) \geq \max(\sigma, 1 - \theta \cdot \gamma_{max})$, where $ \gamma_{max} = \max \limits_{j=1}^n (\gamma_j); \gamma_j > 0$, $\sigma$ is the minimum similarity threshold, and $\theta$ is the maximum hyperbox size parameter. It is noted that the size of all hyperbox candidates must be below $\theta$.
\end{lemma}
The proof of Lemma \ref{lemma2} can be found in the \ref{appendix-b}. By using Lemma \ref{lemma2}, we can change the steps of the AGGLO-SM algorithm in Algorithm \ref{alg-agglofullgfmm} to Algorithm \ref{alg-accelagglosm}, and modify the AGGLO-2 algorithm as shown in Algorithm \ref{alg-accelagglo2}. The only change in these accelerated algorithms compared to their original versions is the limitation of pairs of candidates considered during the learning process by a stricter lower bound. The remaining operations are kept unchanged as described in the original algorithms. 

\begin{algorithm} [!ht]
	\caption{The accelerated AGGLO-SM algorithm} \label{alg-accelagglosm}
	\scriptsize{
	\begin{algorithmic} [1]
	    \STATE Lines 1-5 from the \textbf{Algorithm} \ref{alg-agglofullgfmm}
	    \STATE $ S \leftarrow S \setminus \{ s \in S | s < \max(\sigma, 1 - \theta \cdot \gamma_{max}) \}$
	    \STATE Lines 7-23 from \textbf{Algorithm} \ref{alg-acceiolgfmm}
	\end{algorithmic}
	}
\end{algorithm}

\begin{algorithm} [!ht]
	\caption{The accelerated AGGLO-2 algorithm} \label{alg-accelagglo2}
	\scriptsize{
	\begin{algorithmic} [1]
	    \STATE Lines 1-7 from the \textbf{Algorithm} \ref{alg-agglo2}
	    \STATE $ S \leftarrow S \setminus \{ s \in S | s < \max(\sigma, 1 - \theta \cdot \gamma_{max}) \}$
	    \STATE Lines 9-29 from \textbf{Algorithm} \ref{alg-agglo2}
	\end{algorithmic}
	}
\end{algorithm}

As will be shown in the experimental section, these changes to the agglomerative learning algorithms and the usage of the proposed Lemma \ref{lemma2} led to the acceleration of seven times in the training time of the AGGLO-SM algorithm, while the training time of the AGGLO-2 algorithm is reduced from 25 to 37 times on average depending on the similarity measure and dataset deployed.

\paragraph{Time complexity of the accelerated agglomerative learning algorithms} We can see that the accelerated version of the AGGLO-SM algorithm is only reduced by the number of pairs of hyperbox candidates considered in the aggregation process. Therefore, the complexity of the accelerated AGGLO-SM algorithm is $\mathcal{O}(N^2 \cdot (\overline{\mathcal{Z}}_1 \cdot \overline{\mathcal{Y}} \cdot n + \overline{\mathcal{Y}}^2))$, where $\overline{\mathcal{Z}}_1$ is the average number of pairs of hyperbox candidates, the meaning of the other notations is the same as in the description of the original AGGLO-SM algorithm shown in the subsection \ref{agglo-sec}. Similarly, the complexity of the accelerated AGGLO-2 algorithm is $\mathcal{O}(N \cdot \overline{\mathcal{Z}}_1 \cdot \overline{\mathcal{R}} \cdot n)$ in the worst-case, where $\overline{\mathcal{Z}}_1$ is also the average number of pairs of hyperbox candidates considered during the training process of the accelerated AGGLO-2 algorithm. The other parameters have the same meaning as shown in the subsection \ref{agglo-sec} for the original AGGLO-2 algorithm. The accelerated agglomerative algorithms run faster than the original versions due to $\overline{\mathcal{Z}}_1 < \overline{\mathcal{Z}}$. 

\begin{figure}[!ht]
    \centering
    \includegraphics[width=0.6\textwidth]{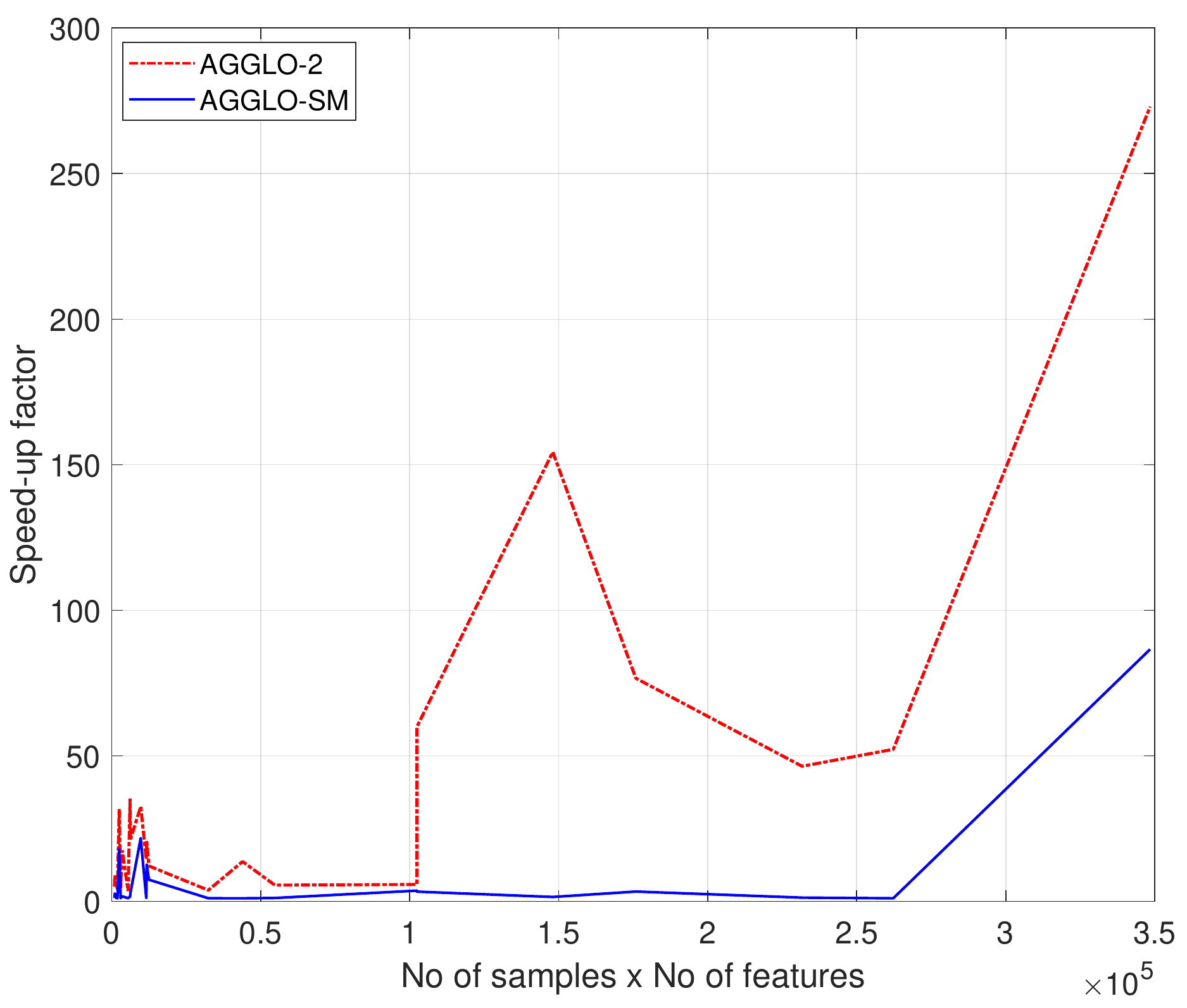}
    \caption{The speedup factor of the AGGLO-2 and AGGLO-SM algorithms according to the number of samples and the number of features over 24 experimental datasets (using the ``longest distance" similarity measure).} \label{fig_speed_up}
\end{figure}

For the time complexity of the AGGLO-2 algorithm, $\overline{\mathcal{Z}}, N, n,$ and $\overline{\mathcal{R}}$ are equally important factors affecting the algorithm time complexity. However, for the complexity of the AGGLO-SM algorithm, the role of $\overline{\mathcal{Z}}$ is much less important than those of $N^2$ and $\overline{\mathcal{Y}}^2$, especially in the large-sized datasets. Therefore, the impact of the reduction of $\overline{\mathcal{Z}}$ on the AGGLO-2 algorithm is more significant compared to the AGGLO-SM algorithm. This fact is confirmed by the experimental results in section \ref{exprt}. As an illustrative example, Fig. \ref{fig_speed_up} shows the speed-up factors of the AGGLO-2 and AGGLO-SM using the proposed method with the ``longest distance" similarity value in correlation to the numbers of samples and features for the 24 experimental datasets. We can see that the acceleration of the AGGLO-2 algorithm is much higher than that of the AGGLO-SM when using the proposed approach. Apart from the numbers of samples and features, the input data distribution and the complexity of the classification problem have a significant effect on the speed-up factor of the proposed approach. For instance, in Table \ref{speedup_agglo2}, two datasets \textit{plant species leaves texture} and \textit{plant species leaves margin} have the same numbers of samples, features, and classes, but their speed-up factors are significantly different.

\section{Experiments} \label{exprt}
\subsection{Experimental datasets and parameter settings}
To evaluate the effectiveness of the proposed method, we conducted the experiments on 24 datasets taken from the UCI machine learning repository \footnote{https://archive.ics.uci.edu/ml/datasets.php}. A summary of these datasets related to the numbers of classes, features, and samples is shown in Table \ref{tbinfo}. 
\begin{table}[!ht]
\caption{The summary of the used datasets}\label{tbinfo}
\centering
\scriptsize{
\begin{tabular}{|c|l|c|c|c|}
\hline
\textbf{ID} & \textbf{Dataset}                & \textbf{\# samples} & \textbf{\# features} & \textbf{\# classes} \\ \hline
1 & blance scale                   & 625                 & 4                    & 3                   \\ \hline
2 & banknote authentication        & 1372                & 4                    & 2                   \\ \hline
3 & blood transfusion              & 748                 & 4                    & 2                   \\ \hline
4 & breast cancer wisconsin       & 699                 & 9                    & 2                   \\ \hline
5 & breast cancer coimbra         & 116                 & 9                    & 2                   \\ \hline
6 & climate model crashes         & 540                 & 18                   & 2                   \\ \hline
7 & connectionist bench sonar     & 208                 & 60                   & 2                   \\ \hline
8 & glass                           & 214                 & 9                    & 6                   \\ \hline
9 & haberman                        & 306                 & 3                    & 2                   \\ \hline
10 & heart                           & 270                 & 13                   & 2                   \\ \hline
11 & ionosphere                      & 351                 & 33                   & 2                   \\ \hline
12 & movement libras                & 360                 & 90                   & 15                  \\ \hline
13 & optical digit                  & 5620                & 62                   & 10                  \\ \hline
14 & page blocks                    & 5473                & 10                   & 2                   \\ \hline
15 & pendigits                       & 10992               & 16                   & 10                  \\ \hline
16 & pima diabetes                  & 768                 & 8                    & 2                   \\ \hline
17 & plant species leaves margin  & 1600                & 64                   & 100                 \\ \hline
18 & plant species leaves texture & 1600                & 64                   & 100                 \\ \hline
19 & ringnorm                        & 7400                & 20                   & 2                   \\ \hline
20 & seeds                           & 210                 & 7                    & 3                   \\ \hline
21 & image segmentation             & 2310                & 19                   & 7                   \\ \hline
22 & spambase                        & 4601                & 57                   & 2                   \\ \hline
23 & spectf heart                   & 267                 & 44                   & 2                   \\ \hline
24 & landsat satellite              & 6435                & 36                   & 6                   \\ \hline
\end{tabular}
}
\end{table}

For each dataset, we carried out 5 times 2-fold cross-validation, and then the average values of the training time and the number of hyperbox candidates considered during the training process are reported in this paper. Experiments were executed on a Intel Xeon Gold 6150 2.7GHz computer with 32GB RAM running Red Hat Enterprise Linux. The algorithms were implemented using Python programming language.

The sensitivity parameter $\gamma_j$ impacts the decreasing speed of the membership function on the $j^{th}$ dimension. If we set a large value of $\gamma_j$, there may be cases that samples are not classified correctly as the membership values for all classes are zero. Therefore, to avoid this situation, we used the sensitivity parameter $\gamma_j = 1; \forall{j \in [1, n]}$ as recommended in \cite{Abe01} when all the input data were normalized to the range of [0, 1]. If the maximum hyperbox size parameter $\theta$ is assigned a large value, the classification accuracy of GFMM is negatively affected. A high classification accuracy is usually achieved for a small value of $\theta$ but it significantly increases the training time \cite{Khuat20comp}. Therefore, to show the efficiency of the proposed method, we used a small value of $\theta = 0.1$ for learning algorithms in this experiment. In the agglomerative learning algorithms, we set the minimum similarity threshold $\sigma = 0$ to assess the impact of the lower bound related to $\theta$ on the training time of algorithms. For $\sigma = 0$, the hyperbox aggregation step depends only on the maximum hyperbox size. We refer the readers to our previous study \cite{Khuat20comp} for the analysis of the impacts of different parameters on the classification performance of GFMM learning algorithms.

The purpose of this experiment is to illustrate and validate the effectiveness of the accelerated learning algorithms in terms of the training time speed up and the reduction of the number of considered hyperbox candidates in comparison to the original versions. The classification accuracy of the accelerated learning algorithms is identical to the original algorithms. The comparison of the learning algorithms of the GFMM model with other state-of-the-art machine learning algorithms can be also found in our previous study \cite{Khuat20comp}.

\subsection{Experimental results for online learning algorithms}
Table \ref{numboxspeeduponln} shows the average speed-up factor and the number of hyperbox candidates considered during the learning process over 10 iterations (5 times 2-fold cross-validation) for each dataset. The speed-up value is computed by dividing the training time of the algorithm without using the lemma by the training time of the algorithm using the lemma to accelerate the learning process. The training time of online learning algorithms are shown in Table \ref{trainingtimeolngfmm} in the \ref{training_time_appendix}.

\begin{table}[!ht]
	\centering
	\caption{Speed-up factor and the number of hyperbox candidates considered during the training process of online learning algorithms}\label{numboxspeeduponln}
	\begin{scriptsize}
		\begin{tabular}{|L{3cm}|C{1.5cm}|C{2cm}|c|C{1.5cm}|C{2cm}|c|}
			\hline
			\multirow{3}{*}{\textbf{Dataset}} & \multicolumn{3}{c|}{\textbf{IOL-GFMM}}                                                                  & \multicolumn{3}{c|}{\textbf{Onln-GFMM}}                                                                 \\ \cline{2-7} 
			& \textbf{Speed-up \break factor} & \multicolumn{2}{l|}{\textbf{Number of hyperbox candidates}} & \textbf{Speed-up factor} & \multicolumn{2}{l|}{\textbf{Number of hyperbox candidates}} \\ \cline{3-4} \cline{6-7} 
			&                                           & \textbf{w/o. lemma}           & \textbf{w. lemma}           &                                           & \textbf{w/o. lemma}           & \textbf{w. lemma}           \\ \hline
			blance scale                      & 5.7227                                    & 20880                         & 0                           & 5.3978                                    & 20880                         & 0                           \\ \hline
			banknote authentication           & 1.3016                                    & 5192.5                        & 914                         & 1.0563                                    & 5129                          & 904                         \\ \hline
			blood transfusion                 & 1.2487                                    & 2154.5                        & 392                         & 1.0568                                    & 1542                          & 320                         \\ \hline
			breast cancer wisconsin           & 3.8746                                    & 14655                         & 0                           & 3.6556                                    & 14655                         & 0                           \\ \hline
			breast cancer coimbra             & 2.3571                                    & 768                           & 2                           & 1.9804                                    & 768                           & 2                           \\ \hline
			climate model crashes             & 6.2478                                    & 30634                         & 0                           & 6.0559                                    & 30634                         & 0                           \\ \hline
			connectionist bench sonar         & 2.8302                                    & 2664.5                        & 0                           & 2.5664                                    & 2664.5                        & 0                           \\ \hline
			glass                             & 1.3178                                    & 567                           & 49.5                        & 1.1009                                    & 567                           & 49.5                        \\ \hline
			haberman                          & 1.3418                                    & 960.5                         & 143.5                       & 1.122                                     & 913.5                         & 141.5                       \\ \hline
			heart                             & 3.5575                                    & 4479.5                        & 1                           & 3.1069                                    & 4479.5                        & 1                           \\ \hline
			ionosphere                        & 2.7678                                    & 5705                          & 46                          & 1.6163                                    & 5705                          & 46                          \\ \hline
			movement libras                   & 1.5575                                    & 676                           & 24.5                        & 1.0989                                    & 676                           & 24.5                        \\ \hline
			optical digit                     & 4.5451                                    & 393452.5                      & 0                           & 3.0049                                    & 393452.5                      & 0                           \\ \hline
			page blocks                       & 1.3019                                    & 21124.5                       & 2928.5                      & 1.0458                                    & 20825.5                       & 2909.5                      \\ \hline
			pendigits                         & 4.2866                                    & 892103                        & 3421.5                      & 1.0656                                    & 892103                        & 3421.5                      \\ \hline
			pima diabetes                     & 4.9452                                    & 26995.5                       & 197.5                       & 1.8977                                    & 26995.5                       & 197.5                       \\ \hline
			plant species leaves margin       & 2.0362                                    & 2800                          & 0                           & 1.3029                                    & 2800                          & 0                           \\ \hline
			plant species leaves texture      & 7.2026                                    & 309308                        & 16                          & 7.417                                     & 309308                        & 16                          \\ \hline
			ringnorm                          & 12.7486                                   & 2986751.5                     & 4108.5                      & 3.0377                                    & 2986751.5                     & 4108.5                      \\ \hline
			seeds                             & 1.7213                                    & 959.5                         & 42.5                        & 1.2                                       & 959.5                         & 42.5                        \\ \hline
			image segmentation                & 1.8082                                    & 26153                         & 1052.5                      & 1.0332                                    & 26153                         & 1052.5                      \\ \hline
			spambase                          & 3.3182                                    & 329619.5                      & 2815.5                      & 1.2982                                    & 329619.5                      & 2815.5                      \\ \hline
			spectf heart                      & 3.8481                                    & 5929.5                        & 0                           & 3.7564                                    & 5929.5                        & 0                           \\ \hline
			landsat satellite                 & 2.76                                      & 349032.5                      & 12759                       & 1.0733                                    & 349029                        & 12752                       \\ \hline
			\textbf{Average ratio with over without lemma}                  & \textbf{3.526963}                         & \multicolumn{2}{c|}{\textbf{0.038444}} & \textbf{2.372788}                         & \multicolumn{2}{c|}{\textbf{0.039792}}         \\ \hline
		\end{tabular}
	\end{scriptsize}
\end{table}

It can be easily observed that the online learning algorithms using the proposed lemma are much faster than ones without deploying the lemma. These figures can be explained based on the number of hyperboxes considered during the training process. We can see that the use of lemma has significantly reduced the unsuitable hyperbox candidates that the original versions have to verify. In several datasets, the number of hyperbox candidates is zero in the case of using the proposed lemma because all existing hyperboxes cannot be extended to cover the new input patterns. It means that the resulting models only contain hyperboxes with one data point. In this case, the speed-up of learning process using the proposed lemma is obvious.

In general, the proposed method contributes to the acceleration of IOL-GFMM algorithm more significantly than the Onln-GFMM. This is because the training time of IOL-GFMM is usually faster than Onln-GFMM algorithm with the small value of $\theta$ ($\theta = 0.1$ in this work) \cite{Khuat2020iol}. Therefore, when the number of candidates considered in the IOL-GFMM reduces, the overlap test operation between the extended hyperbox and the existing hyperboxes is conducted much faster. Meanwhile, the original online learning algorithm needs to check overlap and find the dimension to conduct the contraction for each pair of hyperboxes. These operations occupy most of the computational expense of the Onln-GFMM algorithm, so the obtained speed-up of Onln-GFMM is smaller than that of IOL-GFMM algorithm.

\subsection{Experimental results of agglomerative learning algorithms}

\begin{table}[!ht]
	\centering
	\caption{Speed-up factor of the AGGLO-2 algorithm}\label{speedup_agglo2}
	\begin{scriptsize}
		\begin{tabular}{|l|c|c|c|c|}
			\hline
			\textbf{Dataset}             & \textbf{Longest distance} & \textbf{Shortest distance} & \textbf{Mid-max distance} & \textbf{Mid-min distance} \\ \hline
			blance scale                 & 31.7626                   & 31.9529                    & 21.156                    & 21.2067                   \\ \hline
			banknote authentication      & 3.5408                    & 2.6579                     & 2.6302                    & 2.7469                    \\ \hline
			blood transfusion            & 4.3349                    & 2.7389                     & 2.7078                    & 3.2302                    \\ \hline
			breast cancer wisconsin      & 21.3396                   & 21.1669                    & 14.7005                   & 14.5726                   \\ \hline
			breast cancer coimbra        & 8.9                       & 9.0253                     & 6.0635                    & 6.0397                    \\ \hline
			climate model crashes        & 32.5457                   & 32.2009                    & 20.9304                   & 20.89                     \\ \hline
			connectionist bench sonar    & 12.1624                   & 12.0424                    & 7.829                     & 7.699                     \\ \hline
			glass                        & 4.1902                    & 3.6753                     & 2.9623                    & 3.1639                    \\ \hline
			haberman                     & 5.0756                    & 3.5084                     & 3.1492                    & 3.7729                    \\ \hline
			heart                        & 17.5549                   & 17.5233                    & 11.4509                   & 11.3709                   \\ \hline
			ionosphere                   & 14.5455                   & 12.7882                    & 8.9468                    & 9.6922                    \\ \hline
			movement libras              & 3.84                      & 3.4975                     & 2.8035                    & 2.9697                    \\ \hline
			optical digit                & 272.9711                  & 259.0673                   & 176.7774                  & 183.3464                  \\ \hline
			page blocks                  & 5.5846                    & 3.4114                     & 3.5063                    & 4.1349                    \\ \hline
			pendigits                    & 76.6387                   & 59.2468                    & 51.1295                   & 57.0118                   \\ \hline
			pima diabetes                & 35.3624                   & 28.1336                    & 21.2801                   & 23.7888                   \\ \hline
			plant species leaves margin  & 5.8103                    & 5.725                      & 4.1252                    & 4.1257                    \\ \hline
			plant species leaves texture & 59.9784                   & 60.189                     & 37.6131                   & 37.3868                   \\ \hline
			ringnorm                     & 154.348                   & 110.4401                   & 89.7151                   & 100.4713                  \\ \hline
			seeds                        & 6.2366                    & 5.2175                     & 4.1833                    & 4.6059                    \\ \hline
			image segmentation           & 13.6811                   & 9.7371                     & 8.6332                    & 10.1224                   \\ \hline
			spambase                     & 52.2312                   & 29.2803                    & 28.7548                   & 34.9655                   \\ \hline
			spectf heart                 & 20.6977                   & 20.6945                    & 13.2112                   & 13.239                    \\ \hline
			landsat satellite            & 46.4068                   & 16.1142                    & 20.3232                   & 32.471                    \\ \hline
			\textbf{Average}             & \textbf{37.9058}          & \textbf{31.66811}          & \textbf{23.52427}         & \textbf{25.54268}         \\ \hline
		\end{tabular}
	\end{scriptsize}
\end{table}

This part reports the experimental results of agglomerative learning algorithms with and without using the proposed lemma. Table \ref{speedup_agglo2} presents the speed-up factor of the AGGLO-2 algorithm with the use of the proposed lemma compared to one without using the lemma for four similarity measures. These values are calculated from the average training time shown in Table \ref{trainingtimeagglo2} in the \ref{training_time_appendix}. Table \ref{numbox_agglo2} describes the number of considered hyperbox candidates of the AGGLO-2 algorithm for each similarity measure. Table \ref{speedup_agglosm} shows the speed-up factors of the AGGLO-SM algorithm on the experimental datasets, which are computed from their training time in Table \ref{trainingtimeagglosm} in the \ref{training_time_appendix}. The number of hyperbox candidates considered during the training process of the AGGLO-SM algorithm with four similarity measures is presented in Table \ref{numboxagglosm}.

\begin{table}[!ht]
	\centering
	\caption{The number of hyperboxes considered during the training process of the AGGLO-2 algorithm}\label{numbox_agglo2}
	\begin{scriptsize}
		\begin{tabular}{|L{2.5cm}|c|c|c|c|c|c|c|c|}
			\hline
			\multirow{2}{*}{\textbf{Dataset}} & \multicolumn{2}{c|}{\textbf{Longest distance}} & \multicolumn{2}{c|}{\textbf{Shortest distance}} & \multicolumn{2}{c|}{\textbf{Mid-max distance}} & \multicolumn{2}{c|}{\textbf{Mid-min distance}} \\ \cline{2-9} 
			& \textbf{w/o. lemma}     & \textbf{w. lemma}    & \textbf{w/o. lemma}     & \textbf{w. lemma}     & \textbf{w/o. lemma}     & \textbf{w. lemma}    & \textbf{w/o. lemma}     & \textbf{w. lemma}    \\ \hline
			blance scale                      & 41760                   & 0                    & 41760                   & 0                     & 41760                   & 0                    & 41760                   & 0                    \\ \hline
			banknote authentication           & 22340                   & 552                  & 22105.5                 & 2896                  & 22300                   & 1675.5               & 22285.5                 & 1046                 \\ \hline
			blood transfusion                 & 17806.5                 & 447                  & 14332                   & 1952.5                & 14470.5                 & 1256.5               & 17637.5                 & 836.5                \\ \hline
			breast cancer wisconsin           & 119974                  & 106.5                & 119974                  & 106.5                 & 119974                  & 106.5                & 119974                  & 106.5                \\ \hline
			breast cancer coimbra             & 3038.5                  & 2                    & 3038.5                  & 2                     & 3038.5                  & 2                    & 3038.5                  & 2                    \\ \hline
			climate model crashes             & 61268                   & 0                    & 61268                   & 0                     & 61268                   & 0                    & 61268                   & 0                    \\ \hline
			connectionist bench sonar         & 5329                    & 0                    & 5329                    & 0                     & 5329                    & 0                    & 5329                    & 0                    \\ \hline
			glass                             & 3004                    & 36.5                 & 2954.5                  & 120.5                 & 2954                    & 107.5                & 3005                    & 60.5                 \\ \hline
			haberman                          & 5912.5                  & 93.5                 & 4913.5                  & 398.5                 & 4905.5                  & 275.5                & 5746.5                  & 142.5                \\ \hline
			heart                             & 13311.5                 & 1                    & 13311.5                 & 1                     & 13311.5                 & 1                    & 13311.5                 & 1                    \\ \hline
			ionosphere                        & 33598                   & 27                   & 26748                   & 167.5                 & 26747.5                 & 134.5                & 33599                   & 36                   \\ \hline
			movement libras                   & 3897                    & 34.5                 & 3127.5                  & 55                    & 3127.5                  & 54                   & 3897                    & 36                   \\ \hline
			optical digit                     & 786905                  & 0                    & 786905                  & 0                     & 786905                  & 0                    & 786905                  & 0                    \\ \hline
			page blocks                       & 235607                  & 2522                 & 164723.5                & 15074.5               & 194023                  & 8211.5               & 231764.5                & 5870                 \\ \hline
			pendigits                         & 5080436                 & 1276                 & 5077333                 & 19843                 & 5077333                 & 11493.5              & 5080439                 & 2128.5               \\ \hline
			pima diabetes                     & 128175.5                & 61.5                 & 100825.5                & 576.5                 & 100825.5                & 372.5                & 128175.5                & 84                   \\ \hline
			plant species leaves margin       & 5600                    & 0                    & 5600                    & 0                     & 5600                    & 0                    & 5600                    & 0                    \\ \hline
			plant species leaves texture      & 1232425                 & 12.5                 & 1232425                 & 32.5                  & 1232425                 & 32.5                 & 1232425                 & 12.5                 \\ \hline
			ringnorm                          & 11629867                & 250                  & 11631487                & 29762.5               & 11631487                & 14020.5              & 11629867                & 311                  \\ \hline
			seeds                             & 3333                    & 28.5                 & 3289.5                  & 120                   & 3289.5                  & 95                   & 3333                    & 34.5                 \\ \hline
			image segmentation                & 181993                  & 570                  & 157079.5                & 4098.5                & 157231                  & 2709                 & 181772                  & 1173                 \\ \hline
			spambase                          & 3707848                 & 1150                 & 3340482                 & 51622.5               & 3345250                 & 21569.5              & 3707269                 & 4208.5               \\ \hline
			spectf heart                      & 11859                   & 0                    & 11859                   & 0                     & 11859                   & 0                    & 11859                   & 0                    \\ \hline
			landsat satellite                 & 2021658                 & 1187.5               & 1956507                 & 81882.5               & 1956399                 & 39069                & 2438703                 & 11493                \\ \hline
			\textbf{Average ratio with over without lemma}                  & \multicolumn{2}{c|}{\textbf{0.004713}}      & \multicolumn{2}{c|}{\textbf{0.026592}} & \multicolumn{2}{c|}{\textbf{0.016686}}    & \multicolumn{2}{c|}{\textbf{0.008346}}     \\ \hline
		\end{tabular}
	\end{scriptsize}
\end{table}

In general, the use of the proposed lemma makes the agglomerative learning algorithm much faster because the number of candidates for the hyperbox aggregation process is considerably reduced. Among two versions of the batch learning algorithms, the influence of the proposed lemma on the performance of the AGGLO-2 is more significant compared to the AGGLO-SM. It is due to the fact that the number of hyperbox candidates for the aggregation process in the original AGGLO-2 is higher than the AGGLO-SM algorithm. For several datasets in the AGGLO-2 algorithm such as \textit{optical digit}, \textit{pendigits}, \textit{rignorm}, and \textit{spambase}, the use of the proposed lemma accelerates the training process from 50 to nearly 280 times compared to the original version. Although the AGGLO-SM algorithm cannot obtain such speed-up, the training time also reduces considerably when using the proposed lemma. These results confirm that our proposed method is efficient for all learning algorithms of the GFMM neural network.

\begin{table}[!ht]
	\centering
	\caption{Speed-up factor of the AGGLO-SM algorithm}\label{speedup_agglosm}
	\begin{scriptsize}
	\begin{tabular}{|l|c|c|c|c|}
		\hline
		\textbf{Dataset}             & \textbf{Longest distance} & \textbf{Shortest distance} & \textbf{Mid-max distance} & \textbf{Mid-min distance} \\ \hline
		blance scale                 & 17.9903                   & 17.9709                    & 17.8937                   & 17.8841                   \\ \hline
		banknote authentication      & 1.1323                    & 1.1549                     & 1.1503                    & 1.1634                    \\ \hline
		blood transfusion            & 1.1521                    & 1.1381                     & 1.1133                    & 1.1198                    \\ \hline
		breast cancer wisconsin      & 3.0204                    & 3.0734                     & 2.9323                    & 3.0311                    \\ \hline
		breast cancer coimbra        & 2.6782                    & 2.6897                     & 2.6782                    & 2.6782                    \\ \hline
		climate model crashes        & 21.6677                   & 21.4537                    & 21.7                      & 21.4056                   \\ \hline
		connectionist bench sonar    & 7.4505                    & 7.4286                     & 7.4505                    & 7.4615                    \\ \hline
		glass                        & 1.0887                    & 1.0678                     & 1.0637                    & 1.0775                    \\ \hline
		haberman                     & 1.1469                    & 1.1208                     & 1.1308                    & 1.1521                    \\ \hline
		heart                        & 1.7236                    & 1.7073                     & 1.6763                    & 1.6911                    \\ \hline
		ionosphere                   & 1.2263                    & 1.3166                     & 1.2595                    & 1.2782                    \\ \hline
		movement libras              & 1.0842                    & 1.0981                     & 1.0841                    & 1.0845                    \\ \hline
		optical digit                & 86.5943                   & 83.8357                    & 75.9623                   & 83.9551                   \\ \hline
		page blocks                  & 1.1686                    & 1.1677                     & 1.1389                    & 1.1638                    \\ \hline
		pendigits                    & 3.3746                    & 1.6839                     & 1.591                     & 3.0278                    \\ \hline
		pima diabetes                & 1.5718                    & 1.4924                     & 1.519                     & 1.5698                    \\ \hline
		plant species leaves margin  & 3.6856                    & 3.7029                     & 3.6874                    & 3.6836                    \\ \hline
		plant species leaves texture & 3.3463                    & 3.3465                     & 3.3581                    & 3.3449                    \\ \hline
		ringnorm                     & 1.5222                    & 1.4626                     & 1.4874                    & 1.5244                    \\ \hline
		seeds                        & 1.1595                    & 1.1351                     & 1.1399                    & 1.1459                    \\ \hline
		image segmentation           & 1.0589                    & 1.0521                     & 1.0545                    & 1.0577                    \\ \hline
		spambase                     & 1.0672                    & 1.0647                     & 1.0658                    & 1.0623                    \\ \hline
		spectf heart                 & 12.6057                   & 12.6098                    & 12.5547                   & 12.5587                   \\ \hline
		landsat satellite            & 1.2885                    & 1.0914                     & 1.0982                    & 1.2314                    \\ \hline
		\textbf{Average}             & \textbf{7.49185}          & \textbf{7.286029}          & \textbf{6.949579}         & \textbf{7.348021}         \\ \hline
	\end{tabular}
\end{scriptsize}
\end{table}

\begin{table}[!ht]
	\centering
	\caption{Number of hyperbox candidates considered during the training process of the AGGLO-SM algorithm}\label{numboxagglosm}
	\begin{scriptsize}
	\begin{tabular}{|L{2.5cm}|c|c|c|c|c|c|c|c|}
		\hline
		\multirow{2}{*}{\textbf{Dataset}} & \multicolumn{2}{c|}{\textbf{Longest distance}} & \multicolumn{2}{c|}{\textbf{Shortest distance}} & \multicolumn{2}{c|}{\textbf{Mid-max distance}} & \multicolumn{2}{c|}{\textbf{Mid-min distance}} \\ \cline{2-9} 
		& \textbf{w/o. lemma}     & \textbf{w. lemma}    & \textbf{w/o. lemma}     & \textbf{w. lemma}     & \textbf{w/o. lemma}     & \textbf{w. lemma}    & \textbf{w/o. lemma}     & \textbf{w. lemma}    \\ \hline
		blance scale                      & 18032                   & 0                    & 18032                   & 0                     & 18032                   & 0                    & 18032                   & 0                    \\ \hline
		banknote authentication           & 5003                    & 552.5                & 21709.5                 & 18143.5               & 10925                   & 6988.5               & 6546.5                  & 2390.5               \\ \hline
		blood transfusion                 & 2426                    & 572                  & 3940                    & 2482.5                & 3078.5                  & 1375                 & 2754                    & 947.5                \\ \hline
		breast cancer wisconsin           & 12731.5                 & 106.5                & 12731.5                 & 106.5                 & 12731.5                 & 106.5                & 12731.5                 & 106.5                \\ \hline
		breast cancer coimbra             & 767.5                   & 2                    & 767.5                   & 2                     & 767.5                   & 2                    & 767.5                   & 2                    \\ \hline
		climate model crashes             & 30631                   & 0                    & 30631                   & 0                     & 30631                   & 0                    & 30631                   & 0                    \\ \hline
		connectionist bench sonar         & 2652.5                  & 0                    & 2652.5                  & 0                     & 2652.5                  & 0                    & 2652.5                  & 0                    \\ \hline
		glass                             & 543.5                   & 37                   & 679                     & 198.5                 & 654.5                   & 161.5                & 550                     & 56.5                 \\ \hline
		haberman                          & 1018.5                  & 95.5                 & 1276                    & 418.5                 & 1200.5                  & 285                  & 1003.5                  & 119                  \\ \hline
		heart                             & 419                     & 1                    & 419                     & 1                     & 419                     & 1                    & 419                     & 1                    \\ \hline
		ionosphere                        & 4385.5                  & 26.5                 & 4522                    & 219                   & 4480.5                  & 171.5                & 4403                    & 47                   \\ \hline
		movement libras                   & 690                     & 34.5                 & 727.5                   & 77                    & 720.5                   & 69.5                 & 697                     & 42                   \\ \hline
		optical digit                     & 268395.5                & 0                    & 268395.5                & 0                     & 268395.5                & 0                    & 268395.5                & 0                    \\ \hline
		page blocks                       & 27414                   & 3158.5               & 203481                  & 188327.5              & 76313.5                 & 59269.5              & 44376.5                 & 20037                \\ \hline
		pendigits                         & 801368                  & 1266                 & 825055.5                & 32153                 & 809903                  & 17248.5              & 801416.5                & 2085                 \\ \hline
		pima diabetes                     & 27268.5                 & 60                   & 30319.5                 & 3535.5                & 29136.5                 & 2133.5               & 27328.5                 & 133                  \\ \hline
		plant species leaves margin       & 2792.5                  & 0                    & 2792.5                  & 0                     & 2792.5                  & 0                    & 2792.5                  & 0                    \\ \hline
		plant species leaves texture      & 300635.5                & 12.5                 & 301010                  & 45.5                  & 301010                  & 45.5                 & 300636.5                & 12.5                 \\ \hline
		ringnorm                          & 3000191                 & 241                  & 3952685                 & 965343.5              & 3451147                 & 456658               & 3002038                 & 2119.5               \\ \hline
		seeds                             & 946                     & 29.5                 & 1152                    & 268.5                 & 1080.5                  & 183.5                & 948.5                   & 33                   \\ \hline
		image segmentation                & 25541                   & 571.5                & 68281.5                 & 44145.5               & 49774                   & 25671                & 28740                   & 4101                 \\ \hline
		spambase                          & 328946.5                & 1158                 & 665797.5                & 356241.5              & 513685                  & 198932               & 342728                  & 18653.5              \\ \hline
		spectf heart                      & 5916                    & 0                    & 5916                    & 0                     & 5916                    & 0                    & 5916                    & 0                    \\ \hline
		landsat satellite                 & 384186.5                & 1179.5               & 1592177                 & 1258323               & 1070981                 & 745576               & 446447.5                & 66231                \\ \hline
		\textbf{Average ratio with over without lemma}                  & \multicolumn{2}{c|}{\textbf{0.03153}}      & \multicolumn{2}{c|}{\textbf{0.241009}} & \multicolumn{2}{c|}{\textbf{0.187135}}    & \multicolumn{2}{c|}{\textbf{0.077284}}    \\ \hline
	\end{tabular}
\end{scriptsize}
\end{table}

In the AGGLO-2 algorithm, among four similarity measures, the proposed lemma has the most impact on the longest distance measure and the least influence on the mid-max distance-based similarity measure. This is because the number of hyperbox candidates considered during the original training process using the longest distance-based measure is highest, but after using the proposed method, the number of considered candidates is smallest. Although the number of hyperbox candidates considered in the training process with regard to the middle distance-based similarity measures using the proposed lemma is lower than that of the shortest distance-based measure, its speed-up factor on average is still lower compared to the value of the shortest-based measure. This is due to the fact that the middle-based measures are asymmetrical values, so for each pair of candidates $B_i$ and $B_k$, the training step has to spend time computing both similarity values $s_{ik}$ and $s_{ki}$. The repetition of similarity computation increases the training time and reduces the speed-up factor though there is a reduction in the number of considered candidates.

Similarly, the speed-up factor of the training process using the proposed lemma with regard to the longest distance-based similarity measure in the AGGLO-SM is highest because the number of hyperbox candidates after using the proposed lemma is much lower than those of other similarity measures. Hence, its training time is fastest on average. The influence of the proposed lemma on the AGGLO-SM using mid-max distance measure is still smallest among four similarity measures since its training process has to calculate the similarity values for each pair of hyperboxes twice and the number of hyperbox candidates is relatively high after using the proposed lemma. In contrast to the outcomes of the AGGLO-2, the impact of the proposed lemma on the training time of the AGGLO-SM algorithm using the mid-min distance-based measure is ranked in the second place as the number of candidates is much smaller than those of the shortest and mid-max distance-based measures.

\section{Conclusion} \label{conclu}
One of the drawbacks in the current learning algorithms of the general fuzzy min-max neural network is the consideration of too many candidates in the expansion or aggregation process of hyperboxes. Therefore, this paper presented and proved stricter lower bounds for online and agglomerative learning algorithms of the GFMM neural network. The proposed method reduces significantly the unsuitable hyperbox candidates considered during the learning process, espescially in the AGGLO-2 algorithm. Therefore, the training operations are accelerated when applying our proposed solutions. Experimental results on many datasets confirmed the effectiveness of our approach. In particular, the acceleration factors of the online learning algorithms are from two to three on average, while the training time of the AGGLO-SM algorithm is reduced about seven times on average. Especially, the speed-up factor in the AGGLO-2 algorithm using the proposed lemma can achieve from 30 to 250 on several datasets when the number of unsuitable hyperbox candidates is considerably reduced.

\section*{Declaration of Competing Interest}

The authors declare that they have no known competing financial interests or personal relationships that could have appeared to influence the work reported in this paper.  

\section*{CRediT authorship contribution statement} 

\textbf{Thanh Tung Khuat:} Conceptualization, Methodology, Validation, Software, Writing - original draft. \textbf{Bogdan Gabrys:} Conceptualization, Methodology, Writing - review \& editing, Supervision, Project administration.

\section*{Acknowledgement}
Thanh Tung Khuat would like to thank the UTS-FEIT for awarding him Ph.D. scholarships. The authors would like to thank all anonymous reviewers because of their valuable comments for the quality improvement of this paper.

\appendix

\section{Proof of Lemma 1} \label{appendix-a}
This is a proof of Lemma 1.

\begin{proof}
We need to prove that if the membership degree $b_h(X)$ is below $1-\theta \cdot \gamma_{max}$ then the maximum hyperbox size condition is not satisfied for at least one of the dimensions. First of all, we need prove that if the membership value for the $j^{th}$ dimension $b_h(x_j) < 1 - \theta \cdot \gamma_j$, then the maximum hyperbox size condition is not met for the $j^{th}$ dimension, i.e., $w_j^{new} - v_j^{new} > \theta$. An assumption for this lemma is that all the dimensions of the input hyperbox $ X = [X^l, X^u] $ must satisfy the maximum hyperbox size condition $x_j^u - x_j^l \leq \theta$ and $x_j^u, \; x_j^l \in [0, 1] \; \forall{j} \in [1, n]$. For each $j^{th}$ dimension, there are six cases concerning the positions of the input pattern $ X = [X^l, X^u] $ and the hyperbox $ h = [V, W] $ as follows:

\textbf{\textit{Case 1}}: $ x_j^l \leq v_j \leq x_j^u \leq w_j$. The membership value along the $j^{th}$ dimension is:
$$b_{hj} = b_h(x_j) = \min([1 - f(x_j^u - w_j, \gamma_j)], [1- f(v_j - x_j^l, \gamma_j)]) = 1 - \min((v_j - x_j^l) \cdot \gamma_j, 1)$$

We only consider $ (v_j - x_j^l) \cdot \gamma_j \leq 1$, because in case of $ (v_j - x_j^l) \cdot \gamma_j > 1 \Rightarrow b_{hj} = 0$ and $ 1 < (v_j - x_j^l) \cdot \gamma_j \leq (x_j^u - x_j^l) \cdot \gamma_j \leq \theta \cdot \gamma_j$, thus $1 - \theta \cdot \gamma_j < 0 \Rightarrow b_{hj} = 0 > 1 - \theta \cdot \gamma_j$. Therefore, if $ (v_j - x_j^l) \cdot \gamma_j > 1$, the case $b_{hj} < 1 - \theta \cdot \gamma_j$ will never occur. For $ (v_j - x_j^l) \cdot \gamma_j \leq 1$, we have:
\begin{equation*}
    b_{hj} = 1 - (v_j - x_j^l) \cdot \gamma_j
\end{equation*}

If the hyperbox $ h $ is expanded, then:
\begin{equation*}
    v_j^{new} = \min(v_j, x_j^l) = x_j^l; \quad w_j^{new} = \max(w_j, x_j^u) = w_j
\end{equation*}

We obtain:
\begin{align*}
b_{hj} < 1 - \theta \cdot \gamma_j &\Rightarrow 1 - (v_j - x_j^l) \cdot \gamma_j < 1 - \theta \cdot \gamma_j \Rightarrow v_j - x_j^l > \theta \mbox{ (because of } \gamma_j > 0)\\
&\Rightarrow w_j - x_j^l > \theta \mbox{ (because of } w_j \geq v_j) \Rightarrow w_j^{new} - v_j^{new} > \theta
\end{align*}
Hence, if $b_{hj} < 1 - \theta \cdot \gamma_j$, then $w_j^{new} - v_j^{new} > \theta$ in this case.

\textbf{\textit{Case 2}}: $ v_j \leq x_j^l \leq w_j \leq x_j^u$. The membership value along the $j^{th}$ dimension is:
\begin{equation*}
    b_{hj} = b_h(x_j) = \min([1 - f(x_j^u - w_j, \gamma_j)], [1- f(v_j - x_j^l, \gamma_j)]) = 1 - \min((x_j^u - w_j) \cdot \gamma_j, 1)
\end{equation*}

Similarly to case 1, we only consider $ (x_j^u - w_j) \cdot \gamma_j \leq 1$, thus we have:
\begin{equation*}
    b_{hj} = 1 - (x_j^u - w_j) \cdot \gamma_j
\end{equation*}

If the hyperbox $ h $ is expanded, then:
\begin{equation*}
    v_j^{new} = \min(v_j, x_j^l) = v_j; \quad w_j^{new} = \max(w_j, x_j^u) = x_j^u
\end{equation*}

We have:
	\begin{align*}
	b_{hj} < 1 - \theta \cdot \gamma_j &\Rightarrow 1 - (x_j^u - w_j) \cdot \gamma_j < 1 - \theta \cdot \gamma_j \Rightarrow x_j^u - w_j > \theta \mbox{ (because of } \gamma_j > 0)\\
	&\Rightarrow x_j^u - v_j > \theta \mbox{ (because of } v_j \leq w_j) \Rightarrow w_j^{new} - v_j^{new} > \theta
	\end{align*}
	Hence, if $b_{hj} < 1 - \theta \cdot \gamma_j$, then $w_j^{new} - v_j^{new} > \theta$ in this case.

\textbf{\textit{Case 3}}: $ x_j^l \leq v_j \leq w_j \leq x_j^u$. The membership value along the $j^{th}$ dimension is:
\begin{equation*}
    b_{hj} = b_h(x_j) = \min([1 - f(x_j^u - w_j, \gamma_j)], [1- f(v_j - x_j^l, \gamma_j)]) = \min([1 - (x_j^u - w_j) \cdot \gamma_j], [1 - (v_j - x_j^l) \cdot \gamma_j])
\end{equation*}

According to the assumption of the lemma with regard to the input hyperbox, $x_j^u - x_j^l \leq \theta$, thus, $x_j^u - w_j \leq x_j^u - x_j^l \leq \theta$ and $v_j - x_j^l \leq x_j^u - x_j^l \leq \theta$. These lead to $(x_j^u - w_j) \cdot \gamma_j \leq \theta \cdot \gamma_j$ and  $(v_j - x_j^l) \cdot \gamma_j \leq \theta \cdot \gamma_j$ (because of $\gamma_j > 0$) $\Rightarrow 1 - (x_j^u - w_j) \cdot \gamma_j \geq 1 - \theta \cdot \gamma_j \mbox{ and } 1 - (v_j - x_j^l) \cdot \gamma_j \geq 1 - \theta \cdot \gamma_j$. Therefore, $b_{hj} \geq 1 - \theta \cdot \gamma_j$. As a result, in this case, $b_{hj} < 1 - \theta \cdot \gamma_j$ never happens.

\textbf{\textit{Case 4}}: $ v_j \leq x_j^l \leq x_j^u \leq w_j $, we have the membership value: $b_{hj} = 1 \geq 1 - \theta \cdot \gamma_j$, and so $b_{hj} < 1 - \theta \cdot \gamma_j$ never occurs in this case.

\textbf{\textit{Case 5}}: $ v_j \leq w_j \leq x_j^l \leq x_j^u $. The membership value along the $j^{th}$ dimension is:
\begin{equation*}
    b_{hj} = b_h(x_j) = \min([1 - f(x_j^u - w_j, \gamma_j)], [1- f(v_j - x_j^l, \gamma_j)]) = 1 - \min((x_j^u - w_j) \cdot \gamma_j, 1)
\end{equation*}

If the hyperbox $ h $ is expanded, then:
\begin{equation*}
    v_j^{new} = \min(v_j, x_j^l) = v_j; \quad w_j^{new} = \max(w_j, x_j^u) = x_j^u
\end{equation*}

\textbf{\textit{Case 5.1}}: $(x_j^u - w_j) \cdot \gamma_j > 1 \Leftrightarrow x_j^u - w_j > 1/\gamma_j $ due to $\gamma_j > 0$. In addition, $b_{hj} = 0$. We have:
\begin{align*}
b_{hj} < 1 - \theta \cdot \gamma_j &\Rightarrow 0 < 1 - \theta \cdot \gamma_j \Rightarrow \theta \cdot \gamma_j < 1 \Rightarrow \theta < 1/ \gamma_j < x_j^u - w_j \leq x_j^u - v_j \mbox{ (due to $v_j \leq w_j$ and $\gamma_j > 0$)} \\
&\Rightarrow \theta < w_j^{new} - v_j^{new} 
\end{align*}
Hence, if $b_{hj} < 1 - \theta \cdot \gamma_j$, then $w_j^{new} - v_j^{new} > \theta$ in this case.

\textbf{\textit{Case 5.2}}: $(x_j^u - w_j) \cdot \gamma_j \leq 1$, we have:
\begin{equation*}
    b_{hj} = 1 - (x_j^u - w_j) \cdot \gamma_j
\end{equation*}
and
\begin{align*}
    b_{hj} < 1 - \theta \cdot \gamma_j &\Rightarrow 1 - (x_j^u - w_j) \cdot \gamma_j < 1 - \theta \cdot \gamma_j \Rightarrow x_j^u - w_j > \theta \mbox{ (due to $\gamma_j > 0$)} \\
    & \Rightarrow x_j^u - v_j > \theta \mbox{ (because of $v_j \leq w_j$)} \Rightarrow w_j^{new} - v_j^{new} > \theta
\end{align*}
As a result, if $b_{hj} < 1 - \theta \cdot \gamma_j$, then $w_j^{new} - v_j^{new} > \theta$ in this case as well.

\textbf{\textit{Case 6}}: $ x_j^l \leq x_j^u \leq  v_j \leq w_j $. The membership value along the $j^{th}$ dimension is:
\begin{equation*}
    b_{hj} = b_h(x_j) = \min([1 - f(x_j^u - w_j, \gamma_j)], [1- f(v_j - x_j^l, \gamma_j)]) = 1 - \min((v_j - x_j^l) \cdot \gamma_j, 1)
\end{equation*}

If the hyperbox $ h $ is expanded, then:
\begin{equation*}
    v_j^{new} = \min(v_j, x_j^l) = x_j^l; \quad w_j^{new} = \max(w_j, x_j^u) = w_j
\end{equation*}

\textbf{\textit{Case 6.1}}: $(v_j - x_j^l) \cdot \gamma_j > 1 \Leftrightarrow v_j - x_j^l > 1/\gamma_j $ due to $\gamma_j > 0$. In addition, $b_{hj} = 0$. We obtain:
\begin{align*}
b_{hj} < 1 - \theta \cdot \gamma_j &\Rightarrow 0 < 1 - \theta \cdot \gamma_j \Rightarrow \theta \cdot \gamma_j < 1 \Rightarrow \theta < 1/ \gamma_j < v_j - x_j^l \leq w_j - x_j^l\mbox{ (due to $v_j \leq w_j$ and $\gamma_j > 0$)} \\
&\Rightarrow \theta < w_j^{new} - v_j^{new} 
\end{align*}
Hence, if $b_{hj} < 1 - \theta \cdot \gamma_j$, then $w_j^{new} - v_j^{new} > \theta$ in this case.

\textbf{\textit{Case 6.2}}: $(v_j - x_j^l) \cdot \gamma_j \leq 1$, we obtain:
\begin{equation*}
    b_{hj} = 1 - (v_j - x_j^l) \cdot \gamma_j
\end{equation*}
and:
\begin{align*}
	b_{hj} < 1 - \theta \cdot \gamma_j &\Rightarrow 1 - (v_j - x_j^l) \cdot \gamma_j < 1 - \theta \cdot \gamma_j \Rightarrow v_j - x_j^l > \theta \mbox{ (due to $\gamma_j > 0$)} \\
	& \Rightarrow w_j - x_j^l > \theta \mbox{ (because of $v_j \leq w_j$)} \Rightarrow w_j^{new} - v_j^{new} > \theta
\end{align*}
	As a result, if $b_{hj} < 1 - \theta \cdot \gamma_j$, then $w_j^{new} - v_j^{new} > \theta$ in this case as well.

From the six above cases, we can see that for each dimension $j$ if $b_{hj} < 1 - \theta \cdot \gamma_j$, then $w_j^{new} - v_j^{new} > \theta$. 
Given that the membership function for the input hyperbox $X$ is $ b_h(X) = \min \limits_{j = 1}^n b_{hj}$ and $ 0 \leq \theta \cdot \gamma_j \leq \theta \cdot \max \limits_{j = 1}^n \gamma_j = \theta \cdot \gamma_{max} \Rightarrow 1 - \theta \cdot \gamma_j \geq 1 - \theta \cdot \gamma_{max}$. Therefore, if $b_{hj} < 1 - \theta \cdot \gamma_{max}$, then $w_j^{new} - v_j^{new} > \theta$ for every dimension $j$. As a result, if $b_h(X) < 1 - \theta \cdot \gamma_{max}$, then the maximum hyperbox size condition is not satisfied for the expanded hyperbox. The lemma is proved.
\end{proof}

\section{Proof of Lemma 2} \label{appendix-b}
This is a proof of Lemma 2.

\begin{proof}
An underlying assumption in this lemma is that all input hyperboxes and aggregatable hyperbox candidates have sizes less than $\theta$. If not, the aggregation process will not be possible. First of all, we need to prove that if the similarity value $s_{ik} = s(B_i, B_k) < 1 - \theta \cdot \gamma_{max}$, then the maximum hyperbox size constraint is not satisfied for at least one of the dimensions of the hyperbox aggregated from $B_i$ and $B_k$. Therefore, we will prove that if $s_{ik}^j < 1 - \theta \cdot \gamma_j$ for any dimension $ j \in [1, n] $, then the maximum hyperbox size condition is not met for that $j^{th}$ dimension, i.e., $w_j^{new} - v_j^{new} > \theta$. We see that the similarity measure using middle distance between two hyperboxes is the same as the membership value between a hyperbox and an input pattern. Therefore, the proof is the same as in the appendix A. Here, we only prove the above condition for the longest and shortest distance measures.

\subsection*{\textbf{Using the shortest distance based similarity measure}}
The shortest distance based similarity value for the $j^{th}$ dimension is computed as follows:
\begin{equation*}
\widetilde{s}_{ik}^j = \min[1 - f(v_{kj} - w_{ij}, \gamma_j), 1 - f(v_{ij} - w_{kj}, \gamma_j)]
\end{equation*}
    
For each $j^{th}$ dimension, there are six cases concerning the positions of the hyperbox $ B_i = [V_i, W_i] $ and the hyperbox $ B_k = [V_k, W_k] $ as follows:

\textbf{\textit{Case 1}}: $ v_{ij} \leq v_{kj} \leq w_{ij} \leq w_{kj}$. The similarity value: $\widetilde{s}_{ik}^j = 1 \geq 1 - \theta \cdot \gamma_j$ (because of $\theta \cdot \gamma_j > 0$). Therefore, $\widetilde{s}_{ik}^j < 1 - \theta \cdot \gamma_j$ will never happen in this case.

\textbf{\textit{Case 2}}: $ v_{kj} \leq v_{ij} \leq w_{kj} \leq w_{ij}$. The similarity value: $\widetilde{s}_{ik}^j = 1 \geq 1 - \theta \cdot \gamma_j$. Therefore, $\widetilde{s}_{ik}^j < 1 - \theta \cdot \gamma_j$ will never happen in this case as well.

\textbf{\textit{Case 3}}: $ v_{kj} \leq w_{kj} \leq v_{ij} \leq w_{ij}$. The coordinate at the $j^{th}$ dimension of the hyperbox aggregated from $B_i$ and $B_k$ is:
\begin{equation*}
    v_j^{new} = \min(v_{ij}, v_{kj}) = v_{kj}; \quad w_j^{new} = \max(w_{ij}, w_{kj}) = w_{ij}
\end{equation*} The similarity value: $\widetilde{s}_{ik}^j = 1 - \min[1, (v_{ij} - w_{kj}) \cdot \gamma_j ]$

\textbf{\textit{Case 3.1}}: $(v_{ij} - w_{kj}) \cdot \gamma_j > 1 \Leftrightarrow v_{ij} - w_{kj} > 1 / \gamma_j $ (because of $\gamma_j > 0$). In this case: $\widetilde{s}_{ik}^j = 0 $. We have:
\begin{align*}
&\widetilde{s}_{ik}^j < 1 - \theta \cdot \gamma_j  \Rightarrow 0 < 1 - \theta \cdot \gamma_j \Rightarrow \theta \cdot \gamma_j < 1 \Rightarrow \theta < 1/ \gamma_j < v_{ij} - w_{kj} \leq w_{ij} - w_{kj} \leq w_{ij} - v_{kj}\\
& \mbox{(due to $v_{kj} \leq w_{kj}, v_{ij} \leq w_{ij}, \gamma_j > 0$)} \Rightarrow \theta < w_j^{new} - v_j^{new}
\end{align*}
Therefore, if $\widetilde{s}_{ik}^j < 1 - \theta \cdot \gamma_j$, then $w_j^{new} - v_j^{new} > \theta$ in this case.

\textbf{\textit{Case 3.2}}: $(v_{ij} - w_{kj}) \cdot \gamma_j \leq 1 \Rightarrow \widetilde{s}_{ik}^j = 1 - (v_{ij} - w_{kj}) \cdot \gamma_j$. We obtain:
\begin{align*}
\widetilde{s}_{ik}^j < 1 - \theta \cdot \gamma_j  & \Rightarrow 1 - (v_{ij} - w_{kj}) \cdot \gamma_j < 1 - \theta \cdot \gamma_j \Rightarrow \theta < v_{ij} - w_{kj} \mbox{ (due to $\gamma_j > 0$)} \\
& \Rightarrow \theta < w_{ij} - w_{kj} \leq w_{ij} - v_{kj} = w_j^{new} - v_j^{new} \mbox{ (because of $v_{kj} \leq w_{kj}; v_{ij} \leq w_{ij}$)}
\end{align*}
Therefore, if $\widetilde{s}_{ik}^j < 1 - \theta \cdot \gamma_j$, then $w_j^{new} - v_j^{new} > \theta$ in this case as well.

\textbf{\textit{Case 4}}: $ v_{ij} \leq w_{ij} \leq v_{kj} \leq w_{kj}$. This case is proved similarly to case 3.

\textbf{\textit{Case 5}}: $ v_{kj} \leq v_{ij} \leq w_{ij} \leq w_{kj}$. The similarity value: $\widetilde{s}_{ik}^j = 1 \geq 1 - \theta \cdot \gamma_j$. Therefore, $\widetilde{s}_{ik}^j < 1 - \theta \cdot \gamma_j$ will never happen in this case.

\textbf{\textit{Case 6}}: $ v_{ij} \leq v_{kj} \leq w_{kj} \leq w_{ij}$. The similarity value: $\widetilde{s}_{ik}^j = 1 \geq 1 - \theta \cdot \gamma_j$. Hence, $\widetilde{s}_{ik}^j < 1 - \theta \cdot \gamma_j$ will never happen in this case as well.

\subsection*{\textbf{Using the longest distance based similarity measure}}
The longest distance based similarity value for the $j^{th}$ is calculated as follows:
\begin{equation*}
\widehat{s}_{ik}^j = \min[1 - f(w_{kj} - v_{ij}, \gamma_j), 1 - f(w_{ij} - v_{kj}, \gamma_j)]
\end{equation*}

For each $j^{th}$ dimension, we also consider in turn six cases relevant to the positions of the hyperbox $ B_i = [V_i, W_i] $ and the hyperbox $ B_k = [V_k, W_k] $ as follows:

\textbf{\textit{Case 1}}: $ v_{ij} \leq v_{kj} \leq w_{ij} \leq w_{kj}$. The coordinate at the $j^{th}$ dimension of the hyperbox aggregated from $B_i$ and $B_k$ is:
\begin{equation*}
    v_j^{new} = \min(v_{ij}, v_{kj}) = v_{ij}; \quad w_j^{new} = \max(w_{ij}, w_{kj}) = w_{kj}
\end{equation*} The similarity value: $\widehat{s}_{ik}^j = \min[1 - \min((w_{kj} - v_{ij}) \cdot \gamma_j, 1), 1 - \min((w_{ij} - v_{kj}) \cdot \gamma_j,1)]$. In this case, we have $w_{kj} - v_{ij} \geq w_{kj} - v_{kj} \geq w_{ij} - v_{kj} \Rightarrow \min((w_{kj} - v_{ij}) \cdot \gamma_j, 1) \geq \min((w_{ij} - v_{kj}) \cdot \gamma_j,1)$. Therefore, $\widehat{s}_{ik}^j = 1 - \min((w_{kj} - v_{ij}) \cdot \gamma_j, 1)$

\textbf{\textit{Case 1.1}}: $(w_{kj} - v_{ij}) \cdot \gamma_j > 1 \Leftrightarrow w_{kj} - v_{ij} > 1 / \gamma_j $ (because of $\gamma_j > 0$). In this case: $\widehat{s}_{ik}^j = 0 $. We have:
\begin{align*}
\widehat{s}_{ik}^j < 1 - \theta \cdot \gamma_j \Rightarrow 0 < 1 - \theta \cdot \gamma_j \Rightarrow \theta < 1/ \gamma_j \mbox{ (due to $\gamma_j > 0$)} \Rightarrow \theta < w_{kj} - v_{ij} = w_j^{new} - v_j^{new}
\end{align*} 
Therefore, if $\widehat{s}_{ik}^j < 1 - \theta \cdot \gamma_j$, then $w_j^{new} - v_j^{new} > \theta$.

\textbf{\textit{Case 1.2}}: $(w_{kj} - v_{ij}) \cdot \gamma_j \leq 1 \Rightarrow \widehat{s}_{ik}^j = 1 - (w_{kj} - v_{ij}) \cdot \gamma_j $. We have:
	\begin{align*}
	\widehat{s}_{ik}^j < 1 - \theta \cdot \gamma_j \Rightarrow 1 - (w_{kj} - v_{ij}) \cdot \gamma_j < 1 - \theta \cdot \gamma_j \Rightarrow \theta < w_{kj} - v_{ij} \mbox{ (due to $\gamma_j > 0$)} \Rightarrow \theta < w_j^{new} - v_j^{new}
	\end{align*} 
	Therefore, if $\widehat{s}_{ik}^j < 1 - \theta \cdot \gamma_j$, then $w_j^{new} - v_j^{new} > \theta$ in this case.

\textbf{\textit{Case 2}}: $ v_{kj} \leq v_{ij} \leq w_{kj} \leq w_{ij}$. This case is proved similarity to case 1.

\textbf{\textit{Case 3}}: $ v_{kj} \leq w_{kj} \leq v_{ij} \leq w_{ij}$. The coordinate at the $j^{th}$ dimension of the hyperbox aggregated from $B_i$ and $B_k$ is:
\begin{equation*}
    v_j^{new} = \min(v_{ij}, v_{kj}) = v_{kj}; \quad w_j^{new} = \max(w_{ij}, w_{kj}) = w_{ij}
\end{equation*} The similarity value: $\widehat{s}_{ik}^j = 1 - \min((w_{ij} - v_{kj}) \cdot \gamma_j,1)$.

\textbf{\textit{Case 3.1}}: $(w_{ij} - v_{kj}) \cdot \gamma_j > 1 \Leftrightarrow w_{ij} - v_{kj} > 1 / \gamma_j $ (because of $\gamma_j > 0$). In this case: $\widehat{s}_{ik}^j = 0 $. We have:
	\begin{align*}
	\widehat{s}_{ik}^j < 1 - \theta \cdot \gamma_j \Rightarrow 0 < 1 - \theta \cdot \gamma_j \Rightarrow \theta < 1/ \gamma_j \mbox{ (due to $\gamma_j > 0$)} \Rightarrow \theta < w_{ij} - v_{kj} = w_j^{new} - v_j^{new}
	\end{align*} 
	Therefore, if $\widehat{s}_{ik}^j < 1 - \theta \cdot \gamma_j$, then $w_j^{new} - v_j^{new} > \theta$ in this case.

\textbf{\textit{Case 3.2}}: $(w_{ij} - v_{kj}) \cdot \gamma_j \leq 1 \Rightarrow \widehat{s}_{ik}^j = 1 - (w_{ij} - v_{kj}) \cdot \gamma_j $. We obtain:
	\begin{align*}
	\widehat{s}_{ik}^j < 1 - \theta \cdot \gamma_j \Rightarrow 1 - (w_{ij} - v_{kj}) \cdot \gamma_j < 1 - \theta \cdot \gamma_j \Rightarrow \theta < w_{ij} - v_{kj} \mbox{ (due to $\gamma_j > 0$)} \Rightarrow \theta < w_j^{new} - v_j^{new}
	\end{align*} 
	Therefore, if $\widehat{s}_{ik}^j < 1 - \theta \cdot \gamma_j$, then $w_j^{new} - v_j^{new} > \theta$ in this case.

\textbf{\textit{Case 4}}: $ v_{ij} \leq w_{ij} \leq v_{kj} \leq w_{kj}$. This case is proved similarly to case 3.

\textbf{\textit{Case 5}}: $ v_{kj} \leq v_{ij} \leq w_{ij} \leq w_{kj}$. The coordinate at the $j^{th}$ dimension of the hyperbox aggregated from $B_i$ and $B_k$ is:
\begin{equation*}
    v_j^{new} = \min(v_{ij}, v_{kj}) = v_{kj}; \quad w_j^{new} = \max(w_{ij}, w_{kj}) = w_{kj}
\end{equation*} The similarity value: $\widehat{s}_{ik}^j = \min[1 - \min((w_{kj} - v_{ij}) \cdot \gamma_j, 1), 1 - \min((w_{ij} - v_{kj}) \cdot \gamma_j,1)]$.

According to the assumption of the lemma the sizes of the input hyperboxes for the aggregation process must be below $\theta$ along each of their $n$ dimensions. Therefore, in this case:
\begin{align*}
 w_{kj} - v_{kj} \leq \theta &\Rightarrow 0 \leq w_{kj} - v_{ij} \leq \theta \mbox{ and } 0 \leq w_{ij} - v_{kj} \leq \theta \mbox{ (because of $v_{kj} \leq v_{ij}$ and $w_{ij} \leq w_{kj}$)} \\
 & \Rightarrow  (w_{kj} - v_{ij}) \cdot \gamma_j \leq \theta \cdot \gamma_j \mbox{ and } (w_{ij} - v_{kj}) \cdot \gamma_j \leq \theta \cdot \gamma_j \mbox{ (because of $\gamma_j > 0$)} \\
 & \Rightarrow \min((w_{kj} - v_{ij}) \cdot \gamma_j, 1) \leq \min(\theta \cdot \gamma_j, 1) \mbox{ and } \min((w_{ij} - v_{kj}) \cdot \gamma_j, 1) \leq \min(\theta \cdot \gamma_j, 1) \\
 & \Rightarrow 1 - \min((w_{kj} - v_{ij}) \cdot \gamma_j, 1) \geq 1 - \min(\theta \cdot \gamma_j, 1) \mbox{ and } 1 - \min((w_{ij} - v_{kj}) \cdot \gamma_j, 1) \geq 1 - \min(\theta \cdot \gamma_j, 1) \\
 & \Rightarrow \widehat{s}_{ik}^j \geq 1 - \min(\theta \cdot \gamma_j, 1) \geq 1 - \theta \cdot \gamma_j
\end{align*}
Therefore, the input hyperboxes size assumption always leads to $\widehat{s}_{ik}^j \geq 1 - \theta \cdot \gamma_j$. As a result, $\widehat{s}_{ik}^j < 1 - \theta \cdot \gamma_j$ will never occur in this case. 		

\textbf{\textit{Case 6}}: $ v_{ij} \leq v_{kj} \leq w_{kj} \leq w_{ij}$. This case is proved similarly to case 5.

From the above proofs, we can see that if the similarity value $s_{ik}^j < 1 - \theta \cdot \gamma_j; \; \forall{j \in [1, n]}$, then the hyperbox aggregated from two hyperboxes $B_i$ and $B_k$ does not satisfy the maximum hyperbox size condition on the $j^{th}$ dimension. We also have $0 \leq \theta \cdot \gamma_j \leq \theta \cdot \gamma_{max} \Rightarrow 1 - \theta \cdot \gamma_j \geq 1 - \theta \cdot \gamma_{max}; \forall{j \in [1, n]}$. Therefore, if the similarity score between two hyperboxes $s_{ik} < 1 - \theta \cdot \gamma_{max}$, then the maximum hyperbox size condition is not satisfied for at least one of the dimensions of the aggregated hyperbox. In addition, in the agglomerative learning, two hyperboxes $B_i$ and $B_k$ are aggregated if their similarity value $s_{ik} \geq \sigma$, where $\sigma$ is a given minimum similarity threshold. From these two conditions, we only need to consider pairs of hyperboxes with similarity values $s_{ik} \geq \max(\sigma, 1 - \theta \cdot \gamma_{max})$ when selecting hyperbox canditates for the aggregation process. Lemma 2 is proved.
\end{proof}

\section{Training time of algorithms}\label{training_time_appendix}
This appendix subsection shows the training time of online and agglomerative learning algorithms from the experiments in this paper. Table \ref{trainingtimeolngfmm} shows the training time of the IOL-GFMM and original online learning algorithms. Table \ref{trainingtimeagglo2} presents the training time of the AGGLO-2 algorithm, while Table \ref{trainingtimeagglosm} decribes the learning time of the AGGLO-SM algorithm.

\begin{table}[!ht]
	\centering
	\caption{Training time of online learning algorithms}\label{trainingtimeolngfmm}
	\begin{scriptsize}
		\begin{tabular}{|l|c|c|c|c|}
			\hline
			\multirow{2}{*}{\textbf{Dataset}} & \multicolumn{2}{c|}{\textbf{IOL-GFMM}}  & \multicolumn{2}{c|}{\textbf{Onln-GFMM}} \\ \cline{2-5} 
			& \textbf{w/o. lemma} & \textbf{w. lemma} & \textbf{w/o. lemma} & \textbf{w. lemma} \\ \hline
			blance scale                      & 0.1465              & 0.0256            & 0.1479              & 0.0274            \\ \hline
			banknote authentication           & 0.0984              & 0.0756            & 0.471               & 0.4459            \\ \hline
			blood transfusion                 & 0.0497              & 0.0398            & 0.1322              & 0.1251            \\ \hline
			breast cancer wisconsin           & 0.1205              & 0.0311            & 0.121               & 0.0331            \\ \hline
			breast cancer coimbra             & 0.0099              & 0.0042            & 0.0101              & 0.0051            \\ \hline
			climate model crashes             & 0.2093              & 0.0335            & 0.2168              & 0.0358            \\ \hline
			connectionist bench sonar         & 0.03                & 0.0106            & 0.029               & 0.0113            \\ \hline
			glass                             & 0.0141              & 0.0107            & 0.0382              & 0.0347            \\ \hline
			haberman                          & 0.0212              & 0.0158            & 0.0469              & 0.0418            \\ \hline
			heart                             & 0.0402              & 0.0113            & 0.0407              & 0.0131            \\ \hline
			ionosphere                        & 0.0584              & 0.0211            & 0.091               & 0.0563            \\ \hline
			movement libras                   & 0.0271              & 0.0174            & 0.0578              & 0.0526            \\ \hline
			optical digit                     & 4.9332              & 1.0854            & 3.7477              & 1.2472            \\ \hline
			page blocks                       & 0.8802              & 0.6761            & 4.8526              & 4.64              \\ \hline
			pendigits                         & 14.4605             & 3.3734            & 179.4647            & 168.4146          \\ \hline
			pima diabetes                     & 0.406               & 0.0821            & 0.6995              & 0.3686            \\ \hline
			plant species leaves margin       & 0.2702              & 0.1327            & 0.1768              & 0.1357            \\ \hline
			plant species leaves texture      & 4.3093              & 0.5983            & 4.5103              & 0.6081            \\ \hline
			ringnorm                          & 38.1094             & 2.9893            & 54.7914             & 18.037            \\ \hline
			seeds                             & 0.0315              & 0.0183            & 0.0714              & 0.0595            \\ \hline
			image segmentation                & 0.7204              & 0.3984            & 10.2108             & 9.8831            \\ \hline
			spambase                          & 5.567               & 1.6777            & 17.4908             & 13.4731           \\ \hline
			spectf heart                      & 0.1039              & 0.027             & 0.1033              & 0.0275            \\ \hline
			landsat satellite                 & 6.4317              & 2.3303            & 58.5758             & 54.5749           \\ \hline
			\textbf{Average}                  & \textbf{3.210358}   & \textbf{0.570238} & \textbf{14.00407}   & \textbf{11.34798} \\ \hline
		\end{tabular}
	\end{scriptsize}
\end{table}

\begin{table}[!ht]
	\centering
	\caption{Training time of the AGGLO-2 algorithm}\label{trainingtimeagglo2}
	\begin{scriptsize}
	\begin{tabular}{|L{2cm}|c|c|c|c|c|c|c|c|}
		\hline
		\multirow{2}{*}{\textbf{Dataset}} & \multicolumn{2}{c|}{\textbf{Longest distance}} & \multicolumn{2}{c|}{\textbf{Shortest distance}} & \multicolumn{2}{c|}{\textbf{Mid-max distance}} & \multicolumn{2}{c|}{\textbf{Mid-min distance}} \\ \cline{2-9} 
		& \textbf{w/o. lemma}     & \textbf{w. lemma}    & \textbf{w/o. lemma}     & \textbf{w. lemma}     & \textbf{w/o. lemma}     & \textbf{w. lemma}    & \textbf{w/o. lemma}     & \textbf{w. lemma}    \\ \hline
		blance scale                      & 0.883                   & 0.0278               & 0.8819                  & 0.0276                & 0.8949                  & 0.0423               & 0.8928                  & 0.0421               \\ \hline
		banknote authentication           & 0.6161                  & 0.174                & 0.6153                  & 0.2315                & 0.6686                  & 0.2542               & 0.6675                  & 0.243                \\ \hline
		blood transfusion                 & 0.4673                  & 0.1078               & 0.3892                  & 0.1421                & 0.417                   & 0.154                & 0.4842                  & 0.1499               \\ \hline
		breast cancer wisconsin           & 2.74                    & 0.1284               & 2.7136                  & 0.1282                & 2.8078                  & 0.191                & 2.7892                  & 0.1914               \\ \hline
		breast cancer coimbra             & 0.0712                  & 0.008                & 0.0713                  & 0.0079                & 0.0764                  & 0.0126               & 0.0761                  & 0.0126               \\ \hline
		climate model crashes             & 1.4255                  & 0.0438               & 1.4104                  & 0.0438                & 1.4442                  & 0.069                & 1.4435                  & 0.0691               \\ \hline
		connectionist bench sonar         & 0.1423                  & 0.0117               & 0.1421                  & 0.0118                & 0.1511                  & 0.0193               & 0.1509                  & 0.0196               \\ \hline
		glass                             & 0.0859                  & 0.0205               & 0.0849                  & 0.0231                & 0.0942                  & 0.0318               & 0.0946                  & 0.0299               \\ \hline
		haberman                          & 0.1477                  & 0.0291               & 0.1249                  & 0.0356                & 0.1351                  & 0.0429               & 0.1562                  & 0.0414               \\ \hline
		heart                             & 0.3037                  & 0.0173               & 0.3014                  & 0.0172                & 0.3149                  & 0.0275               & 0.3127                  & 0.0275               \\ \hline
		ionosphere                        & 0.832                   & 0.0572               & 0.6701                  & 0.0524                & 0.6898                  & 0.0771               & 0.8597                  & 0.0887               \\ \hline
		movement libras                   & 0.1728                  & 0.045                & 0.142                   & 0.0406                & 0.1612                  & 0.0575               & 0.1963                  & 0.0661               \\ \hline
		optical digit                     & 323.1978                & 1.184                & 324.0414                & 1.2508                & 324.5103                & 1.8357               & 331.7654                & 1.8095               \\ \hline
		page blocks                       & 14.5953                 & 2.6135               & 10.8555                 & 3.1821                & 13.2088                 & 3.7672               & 15.3054                 & 3.7015               \\ \hline
		pendigits                         & 597.5753                & 7.7973               & 596.2356                & 10.0636               & 599.6309                & 11.7277              & 599.684                 & 10.5186              \\ \hline
		pima diabetes                     & 6.3228                  & 0.1788               & 4.9712                  & 0.1767                & 5.0455                  & 0.2371               & 6.3992                  & 0.269                \\ \hline
		plant species leaves margin       & 0.6798                  & 0.117                & 0.6807                  & 0.1189                & 0.7446                  & 0.1805               & 0.7451                  & 0.1806               \\ \hline
		plant species leaves texture      & 119.9449                & 1.9998               & 120.5406                & 2.0027                & 122.1148                & 3.2466               & 121.6378                & 3.2535               \\ \hline
		ringnorm                          & 1350.8695               & 8.7521               & 1352.427                & 12.2458               & 1357.874                & 15.1354              & 1354.012                & 13.4766              \\ \hline
		seeds                             & 0.174                   & 0.0279               & 0.1727                  & 0.0331                & 0.1849                  & 0.0442               & 0.187                   & 0.0406               \\ \hline
		image segmentation                & 11.882                  & 0.8685               & 10.3135                 & 1.0592                & 10.6223                 & 1.2304               & 12.1874                 & 1.204                \\ \hline
		spambase                          & 441.0555                & 8.4443               & 396.2303                & 13.5323               & 400.5978                & 13.9315              & 446.2259                & 12.7619              \\ \hline
		spectf heart                      & 0.6437                  & 0.0311               & 0.6436                  & 0.0311                & 0.6632                  & 0.0502               & 0.6646                  & 0.0502               \\ \hline
		landsat satellite                 & 255.4462                & 5.5045               & 242.8039                & 15.0677               & 243.0923                & 11.9613              & 307.8871                & 9.4819               \\ \hline
		\textbf{Average}                  & \textbf{130.4281}       & \textbf{1.591225}    & \textbf{127.811}        & \textbf{2.480242}     & \textbf{128.5894}       & \textbf{2.680292}    & \textbf{133.5344}       & \textbf{2.405383}    \\ \hline
	\end{tabular}
\end{scriptsize}
\end{table}

\begin{table}[!ht]
	\centering
	\caption{Training time of the AGGLO-SM algorithm}\label{trainingtimeagglosm}
	\begin{scriptsize}
	\begin{tabular}{|L{2cm}|c|c|c|c|c|c|c|c|}
		\hline
		\multirow{2}{*}{\textbf{Dataset}} & \multicolumn{2}{c|}{\textbf{Longest distance}} & \multicolumn{2}{c|}{\textbf{Shortest distance}} & \multicolumn{2}{c|}{\textbf{Mid-max distance}} & \multicolumn{2}{c|}{\textbf{Mid-min distance}} \\ \cline{2-9} 
		& \textbf{w/o. lemma}     & \textbf{w. lemma}    & \textbf{w/o. lemma}     & \textbf{w. lemma}     & \textbf{w/o. lemma}     & \textbf{w. lemma}    & \textbf{w/o. lemma}     & \textbf{w. lemma}    \\ \hline
		blance scale                      & 0.3724                  & 0.0207               & 0.3702                  & 0.0206                & 0.3704                  & 0.0207               & 0.3702                  & 0.0207               \\ \hline
		banknote authentication           & 20.4138                 & 18.0293              & 20.9532                 & 18.1422               & 20.5455                 & 17.8608              & 20.6175                 & 17.722               \\ \hline
		blood transfusion                 & 0.5674                  & 0.4925               & 0.8059                  & 0.7081                & 0.7212                  & 0.6478               & 0.6057                  & 0.5409               \\ \hline
		breast cancer wisconsin           & 0.3999                  & 0.1324               & 0.4103                  & 0.1335                & 0.3982                  & 0.1358               & 0.3998                  & 0.1319               \\ \hline
		breast cancer coimbra             & 0.0233                  & 0.0087               & 0.0234                  & 0.0087                & 0.0233                  & 0.0087               & 0.0233                  & 0.0087               \\ \hline
		climate model crashes             & 0.6912                  & 0.0319               & 0.6951                  & 0.0324                & 0.6944                  & 0.032                & 0.6914                  & 0.0323               \\ \hline
		connectionist bench sonar         & 0.0678                  & 0.0091               & 0.0676                  & 0.0091                & 0.0678                  & 0.0091               & 0.0679                  & 0.0091               \\ \hline
		glass                             & 0.1718                  & 0.1578               & 0.1733                  & 0.1623                & 0.1686                  & 0.1585               & 0.1738                  & 0.1613               \\ \hline
		haberman                          & 0.1577                  & 0.1375               & 0.1893                  & 0.1689                & 0.1893                  & 0.1674               & 0.1598                  & 0.1387               \\ \hline
		heart                             & 0.0212                  & 0.0123               & 0.021                   & 0.0123                & 0.0233                  & 0.0139               & 0.0208                  & 0.0123               \\ \hline
		ionosphere                        & 0.4828                  & 0.3937               & 0.5086                  & 0.3863                & 0.4907                  & 0.3896               & 0.4833                  & 0.3781               \\ \hline
		movement libras                   & 0.2215                  & 0.2043               & 0.225                   & 0.2049                & 0.2218                  & 0.2046               & 0.2208                  & 0.2036               \\ \hline
		optical digit                     & 105.8615                & 1.2225               & 103.9479                & 1.2399                & 98.2648                 & 1.2936               & 104.8599                & 1.249                \\ \hline
		page blocks                       & 494.1239                & 422.8161             & 1409.174                & 1206.817              & 1156.35                 & 1015.346             & 511.3471                & 439.3924             \\ \hline
		pendigits                         & 130.6623                & 38.7196              & 267.1413                & 158.6442              & 250.7539                & 157.6118             & 140.068                 & 46.2601              \\ \hline
		pima diabetes                     & 3.7218                  & 2.3679               & 3.9888                  & 2.6727                & 3.932                   & 2.5885               & 3.7252                  & 2.3731               \\ \hline
		plant species leaves margin       & 0.3564                  & 0.0967               & 0.3577                  & 0.0966                & 0.3562                  & 0.0966               & 0.3562                  & 0.0967               \\ \hline
		plant species leaves texture      & 40.4664                 & 12.0927              & 40.5373                 & 12.1132               & 40.6338                 & 12.1003              & 40.4593                 & 12.0958              \\ \hline
		ringnorm                          & 1327.968                & 872.3743             & 1442.132                & 986.0059              & 1382.036                & 929.1679             & 1332.649                & 874.2278             \\ \hline
		seeds                             & 0.3235                  & 0.279                & 0.3286                  & 0.2895                & 0.3259                  & 0.2859               & 0.3197                  & 0.279                \\ \hline
		image segmentation                & 62.3063                 & 58.843               & 72.7162                 & 69.1137               & 71.0587                 & 67.3851              & 63.6872                 & 60.2109              \\ \hline
		spambase                          & 1506.906                & 1412.081             & 1642.779                & 1543.002              & 1621.651                & 1521.519             & 1516.833                & 1427.837             \\ \hline
		spectf heart                      & 0.3101                  & 0.0246               & 0.3102                  & 0.0246                & 0.3101                  & 0.0247               & 0.3102                  & 0.0247               \\ \hline
		landsat satellite                 & 237.7967                & 184.5571             & 965.5685                & 884.7285              & 908.0397                & 826.8151             & 291.1204                & 236.4152             \\ \hline
		\textbf{Average}                  & \textbf{163.9331}       & \textbf{126.046}     & \textbf{248.8927}       & \textbf{203.5307}     & \textbf{231.5678}       & \textbf{189.7451}    & \textbf{167.8987}       & \textbf{129.9926}    \\ \hline
	\end{tabular}
\end{scriptsize}
\end{table}



\bibliographystyle{plain}
\bibliography{ref}







\end{document}